\documentclass{article} % For LaTeX2e
\usepackage{iclr2025_conference,times}

\usepackage{amsmath,amsfonts,bm}

\def\eqref#1{equation~\ref{#1}}
\def\1{\bm{1}}

\DeclareMathAlphabet{\mathsfit}{\encodingdefault}{\sfdefault}{m}{sl}
\SetMathAlphabet{\mathsfit}{bold}{\encodingdefault}{\sfdefault}{bx}{n}

\DeclareMathOperator*{\argmax}{arg\,max}
\DeclareMathOperator*{\argmin}{arg\,min}

\usepackage{hyperref}
\usepackage{url}

\usepackage{subcaption}
\usepackage{amsmath}
\usepackage{amssymb}
\usepackage{cleveref}
\usepackage{booktabs}
\usepackage{multirow}
\newcommand{\ie}{\emph{i.e.}}

\usepackage{algorithm}
\usepackage{algorithmic}

\usepackage{array}
\usepackage{sidecap}

\usepackage{textgreek}
\usepackage{tikz}
\usepackage{wrapfig}

\title{Indirect Gradient Matching for Adversarial Robust Distillation}

\author{
Hongsin Lee\thanks{These authors contributed equally.}\textsuperscript{$*$}, 
Seungju Cho\footnotemark[1]\textsuperscript{$*$}, 
Changick Kim \\
Korea Advanced Institute of Science and Technology (KAIST)\\
{\normalsize \texttt{\{hongsin04, joyga, changick\}@kaist.ac.kr}} \\ 
}

\iclrfinalcopy % Uncomment for camera-ready version, but NOT for submission.
\begin{document}

\maketitle

\begin{abstract}
Adversarial training significantly improves adversarial robustness, but superior performance is primarily attained with large models. 
This substantial performance gap for smaller models has spurred active research into adversarial distillation (AD) to mitigate the difference. 
Existing AD methods leverage the teacher’s logits as a guide.
In contrast to these approaches, we aim to transfer another piece of knowledge from the teacher, the input gradient.
In this paper, we propose a distillation module termed Indirect Gradient Distillation Module (IGDM) that indirectly matches the student’s input gradient with that of the teacher.
Experimental results show that IGDM seamlessly integrates with existing AD methods, significantly enhancing their performance.
Particularly, utilizing IGDM on the CIFAR-100 dataset improves the AutoAttack accuracy from 28.06\% to 30.32\% with the ResNet-18 architecture and from 26.18\% to 29.32\% with the MobileNetV2 architecture when integrated into the SOTA method without additional data augmentation.
%  \keywords{Adversarial Robustness \and Adversarial Distillation \and Gradient Distillation}
\end{abstract}

\section{Introduction}
\label{sec:intro}

Recently, adversarial attacks have revealed the vulnerabilities of deep learning-based models \citep{FGSM, CW_attack, PGD}, raising critical concerns in safety-important applications \citep{safe2, safe1,wang2023does}.
Thus, much research has been done on defense technologies to make deep learning more reliable \citep{2017_jpeg_defense, carmon2019unlabeled, cohen2019certified, xie2019feature, zhang2022adversarial, jin2023randomized}.
Among adversarial defense mechanisms, adversarial training is one of the most effective methods to enhance adversarial robustness \citep{pang2020bag, bai2021recent, wei2023cfa}.
However, there is a significant performance gap between large and small models in adversarial training.
Since light models with less computational complexity are preferred in practical applications, increasing the robustness of light models is necessary.
To address this, adversarial training incorporates distillation methodologies which are commonly employed to boost the performance of light or small models \citep{ard,iad, rslad,adaad}.
% To address this, knowledge distillation emerges as a prevalent technique, commonly employed to boost the performance of light or small models.
% Notably, adversarial training also incorporates distillation methodologies \citep{ard,rslad,iad,adaad}.

% In the teacher-student architecture of knowledge distillation, existing methods primarily utilize the teacher's feature or logits \citep{hinton2015distilling, ji2021show, kim2021feature, yang2022focal}. 
% In adversarial distillation (AD) works, they utilize the logits of teachers as a guide with their adversarial training \citep{ard,iad,rslad,adaad}.
% Recent studies mainly customize the inner maximization problem of adversarial training by conjugating the teacher model when generating adversarial examples\citep{rslad,adaad}.
In the teacher-student architecture of knowledge distillation, prevailing methods leverage the teacher's features or logits \citep{hinton2015distilling, ji2021show, kim2021feature, yang2022focal}. Adversarial distillation (AD) approaches extend this paradigm by incorporating the teacher's logits as a guide within their adversarial training framework \citep{ard, iad, akd}. Recent studies have specifically focused on tailoring the inner maximization problem of adversarial training, particularly by involving the teacher model in the generation of adversarial examples \citep{rslad, adaad}.
% In this paper, we exploit another knowledge of teachers: input gradient, which is highly effective in the outer minimization of adversarial training.
We leverage input gradients, which have been studied in relation to robustness in adversarially trained models \citep{tsipras2018robustness, engstrom2019adversarial, srinivas2023models}.
% Orthogonal to this research direction, we propose matching gradient information of teacher and student which is helpful outer minimization of adversarial training.

% Here, methods that use only the teacher's output as loss are mainly used, rather than the student's own cross entropy loss. This is due to research results showing that a robust teacher has a wider spread of output than a natural model and that learning from this is more effective in adversarial robustness, and all recently proposed methods utilize the teacher's output. In addition, methods to increase performance by further upgrading the method of generating adversarial images for adversarial training have been proposed. However, the gap between teacher and student is still large, and this can be seen as a limitation of only following the teacher's output.

\begin{figure}[t]
\centering
    \includegraphics[width=0.8\textwidth] {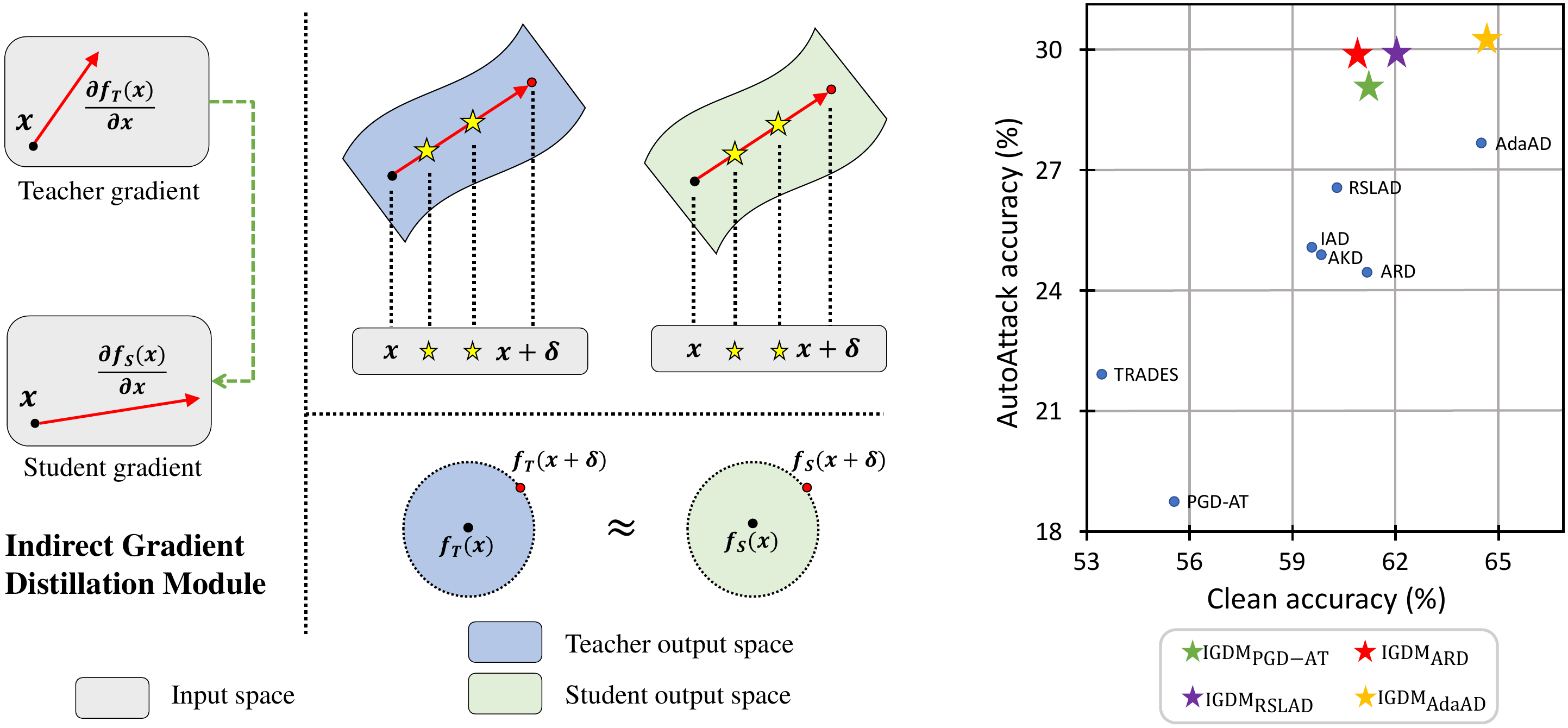}%
    \\
    \vspace{-0.6cm}
    \begin{minipage}[t]{.58\textwidth}
      \subcaption{Conceptual diagram of IGDM}   
       \label{fig:intro_diagram}
    \end{minipage}%
    \begin{minipage}[t]{.37\textwidth}
        \subcaption{Performance of IGDM}
       \label{fig:intro_robustness}%
    \end{minipage}
    \vspace{-0.1cm}
    % \caption{Conceptual diagram of adversarial robust distillation and gradient distillation module. We match the gradient in the input space with knowledge distillation. We name the proposed method the gradient distillation module because the proposed method can be easily combined with existing adversarial robust distillation methods.}
    \caption{ \textbf{(a)} We match the gradient in the input space with knowledge distillation. 
    % If the loss landscape is locally linear, matching the gradient has the effect of matching the points marked with  `stars' around the input, as illustrated in the top right figure.
    Since adversarial robust models are locally linear, matching the gradient has the effect of matching the output points of the surrounding input marked with `stars', as illustrated in the top right figure.
    Consequently, gradient matching involves matching the teacher and student point by point in the output space, as depicted in the bottom right.
    \textbf{(b)} Clean and Autoattack accuracy on ResNet-18 with BDM-AT teacher on CIFAR-100 dataset. IGDM demonstrates significantly improved AutoAttack accuracy compared with other adversarial distillation methods.}
    \label{fig:intro}
\end{figure}

We additionally demonstrate that matching the input gradients between teacher and student contributes to point-wise alignment, which has been studied in state-of-the-art AD methods \citep{adaad}.
% allowing the student to better follow the robust teacher.
In the top right of \Cref{fig:intro_diagram}, if the input gradients between the teacher and the student match as depicted in the red line, then the output of the points located on the red line, denoted as \textit{stars}, will also match.
We show that the output of random points around $x$ also becomes similar to the teacher as illustrated in the bottom right of \Cref{fig:intro_diagram}.
This alignment causes the student to mimic the teacher and reduces the capacity gap between teacher and student.

In this paper, we propose \textit{Indirect Gradient Distillation Module} (IGDM), which indirectly aligns the student's gradient with that of the teacher.
We can easily match input gradients through Taylor expansion, leveraging the locally-linear property of adversarial training.
% One key feature of IGDM is modularity.
Since existing AD methods are mainly designed to match logits, IGDM can be used in conjunction with them.
Through extensive experimental results, we verify that our method successfully complements the robustness of existing AD methods.
We significantly improve existing AD methods as depicted in \Cref{fig:intro_robustness}.
%Additionally, we confirm that adding our loss increases performance for the noise of the larger norm.
%Moreover, when our module is applied to existing adversarial training methods, our module improves defense performance against PGD l1 and PGD l2 attacks, which are unseen norm attacks, and also improves transfer-based attack robustness.
%This means that we distill the teacher's knowledge better and thus improve the performance of students.
Our contributions are as follows:
\begin{itemize}
% \item We devise a way to distill the teacher's gradient information to the student by utilizing the local linearity of adversarial training, in contrast to existing AD methods focusing on distilling logits.
\item We propose a methodology to transfer the gradient information from the teacher to the student through the exploitation of the local linearity inherent in adversarial training.
This stands in contrast to prevailing AD methods, which primarily concentrate on the distillation of logits.
\item Based on the analysis, we propose the Indirect Gradient Distillation Module (IGDM), which indirectly distills the gradient information. Its modular design allows easy integration into existing AD methods.
% \item IGDM notably improves robustness across various attack scenarios, achieving a significant AutoAttack robustness (30.32\%) when trained on the CIFAR-100 dataset using ResNet-18. 
\item We empirically demonstrate that IGDM significantly improves robustness against various attack scenarios, datasets, student models, and teacher models.
%, showcasing notable improvements in AutoAttack robustness.
%trained on the CIFAR-100 dataset: 30.32\% on ResNet-18 and 29.52\% on MobileNetV2.
%Furthermore, our module is effective even without employing the teacher model in the inner maximization.
\end{itemize}

%------------------------------------------------------------------------

\section{Related Work}

\subsection{Adversarial Training}
Adversarial training (AT) involves generating adversarial examples during the training process to improve model robustness.
The adversarial training is a minimization-maximization problem as follows:
\begin{equation}\label{eq:basic_adv}
\begin{split}
    \argmin_{\theta}  \mathbb{E}_{(\mathbf{x}, y)\sim \mathit{D}}\left [    L_{min}\left (f_{\theta}(\mathbf{x}+\boldsymbol{\delta}), y \right ) \right  ], \ 
    \text{where} \ \boldsymbol{\delta} = \argmax_{ \| \boldsymbol{\delta'} \|_p \leq \epsilon}\ L_{max}\left (f_{\theta}(\mathbf{x}+\boldsymbol{\delta'}), y \right ).
\end{split}
\end{equation}
% \begin{equation}\label{eq:basic_adv}
%     \min_{\theta}  \mathbb{E}_{(\mathbf{x}, y)\sim \mathit{D}}\left ( \max_{ \| \boldsymbol{\delta} \|_p \leq \epsilon}\   l_{obj}(f_{\theta}(\mathbf{x}+\boldsymbol{\delta}), y ) \right  ),
% \end{equation}
% \begin{equation}\label{eq:basic_adv}
%     \argmin_{\theta}  \mathbb{E}_{(\mathbf{x}, y)\sim \mathit{D}}\left ( \max_{\boldsymbol{\delta} \in [-\epsilon, \epsilon]} \left  l_{obj}(\mathbf{x}+\boldsymbol{\delta}, y ; f_{\boldsymbol{\theta}}) \right   \right ),
% \end{equation}
Here, $\theta$ is parameters of model $f$, $\mathit{D}$ is a data distribution of $\mathbf{x}$ and corresponding labels $y$, 
$\boldsymbol{\delta}$ is perturbation causing the largest loss within $l_p$ norm of $\epsilon$, $L_{max}$ is an inner maximiazation loss, and $L_{min}$ is an outer minimization loss.
Multi-step PGD attack \citep{PGD} is commonly utilized to solve the inner maximization loss, and various regularization loss functions have been introduced for the outer minimization loss.
TRADES \citep{TRADES} incorporates the Kullback-Leibler (KL) divergence loss between the predictions of clean and adversarial images.
MART\citep{MART} introduces per-sample weights based on the confidence of each sample.
Due to their simplicity and effectiveness, TRADES and MART are commonly used as baseline methods in adversarial training \citep{2020_awp,bai2021improving, jin2022enhancing, tack2022consistency, jin2023randomized,wei2023cfa, qin2019adversarial}. 
Moreover, several strategies such as data augmentation \citep{rebuffi2021fixing, li2023data}, and diverse loss functions \citep{2020_awp}, have been introduced. 

Although defending against strong adversarial attack strategies\citep{AutoAttack, croce2020minimally,andriushchenko2020square} is challenging, highly robust models have also been developed.
Low Temperature Distillation (LTD) \citep{LTD} points out the shortcomings of one-hot labels in adversarial training and advocates the use of soft labels as an alternative approach.
Better Diffusion Models for Adversarial Training (BDM-AT) \citep{wang2023better} explores methods for the more efficient utilization of diffusion models within the context of adversarial training.
Improved Kullback–Leibler Adversarial Training (IKL-AT) \citep{cui2023decoupled} inspects the mechanism of KL divergence loss and proposes the Decoupled Kullback-Leibler divergence loss.
However, given the use of large architectures in these models, there is a necessity to enhance adversarial robustness in smaller models.

\subsection{Adversarial Robust Distillation}

Adversarial distillation (AD) is an effective technique to distill the robustness from large teacher model to small student model \citep{ard, iad, akd, rslad, adaad, kuang2023improving, jung2024peeraid, yin2024adversarial, dong2025adversarially}.
Adversarial Robustness Distillation (ARD) \citep{ard} reveals that students can more effectively acquire robustness when guided by a robust teacher within an adversarial training framework. 
RSLAD \citep{rslad} emphasizes the importance of smooth teacher logits in robust distillation, integrating the teacher's guidance directly into the adversarial image generation process.
Introspective Adversarial Distillation (IAD) \citep{iad} concentrates on assessing the reliability of the teacher and introduces a confidence score of the information provided by the teacher.
AdaAD \citep{adaad} generates more sophisticated adversarial images through the integration of the teacher during the inner maximization process.
Previous AD methods utilize the teacher's logits as a guide, but our approach also incorporates the distillation of gradient information.

\subsection{Gradient Distillation and Input Gradient}
In knowledge distillation,  gradient information has been used in various ways \citep{czarnecki2017sobolev, du2020agree, zhu2021student, lan2023gradient,wang2022gradient}, computed in either input space or weight space.
% These gradients can be computed in either input space or weight space, leading to a distinction between these two domains.
% For instance, when applying knowledge distillation to ensemble models, optimization benefits from incorporating the teacher's weight gradients \citep{du2020agree}. 
For example, an exploration of the diversity among teacher models in the weight gradient space aids in identifying an optimal direction for training the student network \citep{du2020agree}.
Another study examines the capacity gap between teachers and students, focusing on the perspective of weight gradients similarity \citep{zhu2021student}.
Conversely, input space gradients find applications in knowledge distillation for tasks such as classification \citep{czarnecki2017sobolev}, object detection \citep{lan2023gradient}, or language model \citep{wang2022gradient}.
These methods all compute the gradient directly, which is different from our approach.

Additionally, input gradients of adversarially trained models contain semantically meaningful information, aligning with human perception, where image modifications for a specific class resemble human-recognized features \citep{tsipras2018robustness, engstrom2019adversarial, srinivas2023models}. This suggests that robust models capture more human-aligned features, underlining the potential of distilling such valuable insights from robust teacher models to enhance student model performance in adversarial settings.

\section{Method}
In this section, we elaborate on how to distill the input gradients from the robust teacher models. In the analysis of \Cref{sec:direct_matching}, direct gradient matching is shown to be ineffective. Therefore, we propose a novel approach for gradient matching that avoids direct gradient computation.
\subsection{Local Linearity of Adversarial Training}
\label{sec:local_linear}
First, we show that adversarially trained models are capable of first-order Taylor expansion on the input unlike natural training.
% The output for small noise added-input $\textbf{x}$ can be expressed as follows.
For small noise $\boldsymbol{\epsilon}$, the output of $\textbf{x} +\boldsymbol{\epsilon}$ can be expressed as follows.
\begin{wrapfigure}{r}{0.53\textwidth}
    \centering
    \begin{minipage}[c]{.23\textwidth}
      \centering
      \includegraphics[scale=0.18]{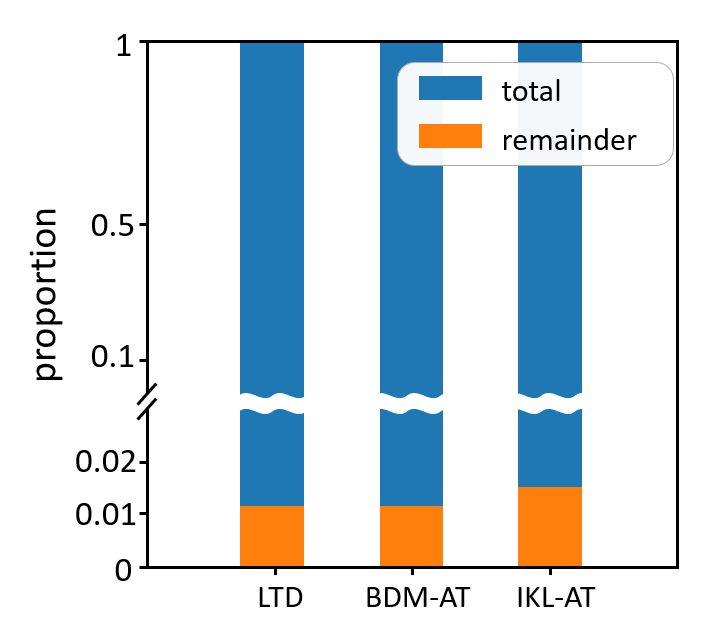}
    \end{minipage}%
    \begin{minipage}[c]{.3\textwidth}
      \centering
      \includegraphics[scale=0.41]{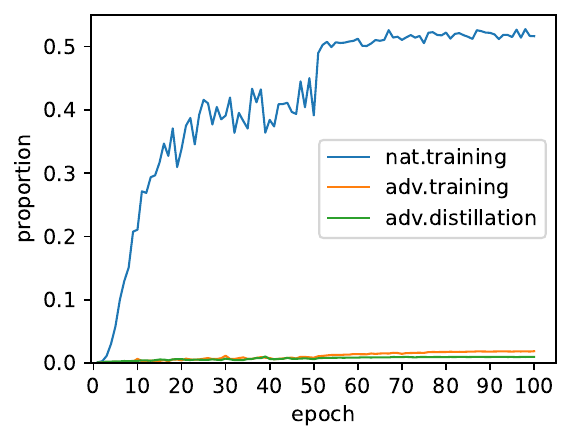}    
    \end{minipage}
      \\
    \vspace{-0.2cm}
    \begin{minipage}[t]{.25\textwidth}
      \subcaption{}   
       \label{fig:ratio_teacher}
    \end{minipage}%
    \begin{minipage}[t]{.25\textwidth}
        \subcaption{}
       \label{fig:ratio_student}%
    \end{minipage}
    \caption{The proportion occupied by the remainder in Taylor expansion. \textbf{(a)} The remainder proportion of an adversarially robust teacher model on CIFAR-100 dataset.} \textbf{(b)} The proportion along with training epochs in natural training, adversarial training, and adversarial distillation using ResNet-18 on CIFAR-100 dataset.%
    \label{fig:ratio}%
    \vspace{-1cm}
\end{wrapfigure}
\begin{equation}\label{eqn:adv_train_local}
    f(\textbf{x} + \boldsymbol{\epsilon}) = \underbrace{f(\textbf{x}) + \left ( \frac{\partial f(\textbf{x})}{\partial \textbf{x}}\right )^T \boldsymbol{\epsilon}}_{\text{first-order approximation}} + \underbrace{\vphantom{ \left ( \frac{\partial f(\textbf{x})}{\partial \textbf{x}}\right)}O(\|\boldsymbol{\epsilon}\|^2)}_{\text{remainder}}.
\end{equation}
Here, $f$ denotes the model, where we use $f$ for brevity instead of   $f_{\theta}$, and $\boldsymbol{\epsilon}$ is sufficiently small noise with the same dimension as the \textbf{x}.
% To investigate the impact of the remainder, we compute the proportion occupied by the remainder in \Cref{eqn:adv_train_local} by applying uniform noise with a size of 8$/$255 as the perturbation $\epsilon$.
To explore the influence of the remainder, we calculate the proportion occupied by the remainder in \Cref{eqn:adv_train_local} by introducing uniform noise with a magnitude of $8/255$ as the perturbation $\epsilon$.
We calculate the remainder proportion as the ratio of the remainder to the total value, \ie, $\frac{\|f(\textbf{x} + \boldsymbol{\epsilon}) - f(\textbf{x}) - \left( \frac{\partial f(\textbf{x})}{\partial \textbf{x}} \right)^T \boldsymbol{\epsilon} \|}{\|f(\textbf{x} + \boldsymbol{\epsilon})\|}$.
We first examine the remainder proportion in an adversarially well-trained model: LTD \citep{LTD}, BDM-AT \citep{wang2023better}, and IKL-AT \citep{cui2023decoupled} where we summarized the performance in \Cref{tab:main_teacher}.
%Then we measure ratio on the entire CIFAR-100 test dataset using ResNet-18 architecture.
% As a result, the ratio of remaining terms $r_2$ was measured to be sufficiently low as in \Cref{fig:ratio_teacher}.
In \Cref{fig:ratio_teacher}, the remainder proportions are computed to be very small, with values of 0.012, 0.012, and 0.016 for three adversarially-trained models LTD, BDM-AT, and IKL-AT, respectively.
% The ratio of the remaining term in the three adversarially-trained models LTD, BDM, and IKL-AT was found to be 0.012, 0.012, and 0.016, respectively.
% In other words, $f(\textbf{x} + \boldsymbol{\epsilon}$ can be approximated
In other words, $f(\textbf{x} + \boldsymbol{\epsilon})$ can be approximated to the first-order Taylor expansion since the remainder proportion is negligible.
% Thus, we conclude that the adversarially well-trained model is locally linear around inputs.
Thus, we utilize this local linearity of the adversarially well-trained models to match input gradients.

Next, we investigate whether the model retains the local linearity in adversarial training, not in the case of a fully-trained model.
We test three training strategies: natural training, adversarial training via ten steps of PGD \citep{PGD}, and adversarial distillation \citep{ard} using the BDM-AT teacher.
In \Cref{fig:ratio_student}, the remainder proportion continuously increases in natural training, while it consistently remains small in both adversarial training and adversarial distillation.
Therefore, employing first-order Taylor expansion is feasible during adversarial training.
In the following section, we demonstrate how the locally linear property enables the matching of input gradients between the teacher and the student.

% \begin{figure*}[t]
% \centering
%     \includegraphics[width=1\textwidth] {figure/model.png}%
%     \caption{Summary diagram of AD with IGDM. 
% IGDM indirectly obtains input gradients of both the teacher and the student by calculating output differences and perform distillation to match the gradients. IGDM can be seamlessly integrated with existing AD methods.}
%     \label{fig:intro2}
% \end{figure*}

\subsection{Gradient Matching via Output Differences}
For an input $\textbf{x}$, we define $\textbf{x}_{\epsilon_1} $ and $\textbf{x}_{\epsilon_2}$ as follows:
\begin{equation}\label{method-1}
    \textbf{x}_{\epsilon_1} = \textbf{x} + \boldsymbol{\epsilon_1}, \ \ \textbf{x}_{\epsilon_2} = \textbf{x} + \boldsymbol{\epsilon_2},
\end{equation}
where $\boldsymbol{\epsilon_1}$ and $\boldsymbol{\epsilon_2}$ represent small random perturbations of the same dimension as the input $\textbf{x}$.
For adversarially trained or training model $f$, the output of $\textbf{x}_{\epsilon_1}$ and $\textbf{x}_{\epsilon_2}$ can be approximated using first-order Taylor expansion through the input space:
\begin{equation}\label{method-2}
\begin{split}
    f(\textbf{x}_{\epsilon_1}) \approx f(\textbf{x}) + 
    \left ( \frac{\partial f(\textbf{x})}{\partial \textbf{x}}\right )^T \boldsymbol{\epsilon_1}, \ \ 
    f(\textbf{x}_{\epsilon_2}) \approx f(\textbf{x}) + 
    \left ( \frac{\partial f(\textbf{x})}{\partial \textbf{x}}\right )^T \boldsymbol{\epsilon_2}, 
\end{split}
\end{equation}
where we neglect the remainder term based on the observations in \Cref{sec:local_linear}.
To extract and align the gradients between the student and the teacher models, we utilize the output differences as follows:
\begin{equation}\label{method-3}
    L = D\left (f_S(\textbf{x}_{\epsilon_1}) - f_S(\textbf{x}_{\epsilon_2}) \ ,\  f_T(\textbf{x}_{\epsilon_1}) - f_T(\textbf{x}_{\epsilon_2}) \right ),
\end{equation}
where $f_S$ and $f_T$ represent the student and teacher models, while $D$ denotes a discrepancy metric like L2 or KL divergence loss.
This loss can be reformulated using \Cref{method-2} as follows:
\begin{equation}
\label{method-4} L = D\left (\left ( \frac{\partial f_S(\textbf{x})}{\partial \textbf{x}}\right )^T (\boldsymbol{\epsilon_1} - \boldsymbol{\epsilon_2}) \ , \ \left ( \frac{\partial f_T(\textbf{x})}{\partial \textbf{x}}\right )^T (\boldsymbol{\epsilon_1} - \boldsymbol{\epsilon_2}) \right ). 
\end{equation} 
If we consider $\boldsymbol{\epsilon_1} - \boldsymbol{\epsilon_2}$ is an arbitrary perturbation vector, varying through the optimization process,  minimizing this loss encourages the alignment of the gradients between the student and the teacher models, meaning \begin{equation}\label{method-5}
\begin{split} \frac{\partial f_S(\textbf{x})}{\partial \textbf{x}} \approx \frac{\partial f_T(\textbf{x})}{\partial \textbf{x}}. 
\end{split} 
\end{equation} 
% While not guaranteeing exact equality, this optimization brings the gradients even closer in response to adversarial perturbations, as discussed in the following section.

%For the distance metric of $D$, we can utilize L2 loss or KL divergence loss.

\subsection{Indirect Gradient Distillation Module (IGDM)}
To effectively integrate the gradient matching through output differences into AD methods, we select the $\boldsymbol{\epsilon_1}$ and $\boldsymbol{\epsilon_2}$ as constant multiples of the adversary perturbation $\boldsymbol{\delta}$ from the AD methods, \ie, $ \boldsymbol{\epsilon_1} - \boldsymbol{\epsilon_2} \propto \boldsymbol{\delta}$.
Since the adversarial perturbations obtained during the training process continuously change, the gradients can be aligned.
Specifically, as the adversarial perturbation $\boldsymbol{\delta}$ employs gradient information, the inner product of gradient and $\boldsymbol{\epsilon_1} - \boldsymbol{\epsilon_2}$ in \Cref{method-4} can be effectively increased with the adversarial perturbation.
In other words, 
$( \frac{\partial f_S(\textbf{x})}{\partial \textbf{x}})^T(\boldsymbol{\epsilon_1} - \boldsymbol{\epsilon_2}) \propto ( \frac{\partial f_S(\textbf{x})}{\partial \textbf{x}})^T \boldsymbol{\delta}$ remains large, facilitating the enhanced alignment of gradients using loss function in \Cref{method-4} during training.
% Furthermore, the adversarial perturbations obtained during the training process continuously change, aligning with the analysis provided in the previous section.
% If $\boldsymbol{\epsilon_1} - \boldsymbol{\epsilon_2}$ were random noise, its inner product would be zero when the noise is orthogonal to the gradient, thereby making no contribution to the alignment of gradients.
%Moreover, as the adversarial perturbation of the origin model has been already generated by the original model for adversarial training, there is no need for further computation.
%Moreover, it has been observed that as the gap between $\boldsymbol{\epsilon_1}$ and $\boldsymbol{\epsilon_2}$ widens to a certain extent, the gradient matching and robustness performance improve even further.

Finally, Indirect Gradient Distillation Module (IGDM) loss is formulated as:
\begin{equation}\label{method6}
\begin{split}
    L_{IGDM} = D\left (f_S(\textbf{x} + \boldsymbol{\delta} ) - f_S(\textbf{x} -  \boldsymbol{\delta} ) 
    \ , \ f_T(\textbf{x} + \boldsymbol{\delta}) - f_T(\textbf{x} -  \boldsymbol{\delta} ) \right ).
\end{split}
\end{equation}
% Here, $T(\alpha)$ is a non-negative, increasing hyperparameter function throughout the training epochs, as applying the IGDM loss at the very early stages is not effective.
% $\beta$ is a non-negative hyperparameter.
%Our model architecture is depicted in \Cref{fig:intro2}.
Since IGDM complements the loss function by matching the gradients between the teacher and student models, the integration of IGDM with other AD methods becomes feasible in the following manner:
\begin{equation}\label{method7}
    L_{min} = L_{AD} + \alpha \cdot L_{IGDM},
\end{equation}
where $\alpha$ is a hyperparameter and $L_{AD}$ stands for the outer minimization loss of other AD methods such as ARD \citep{ard}, RSLAD \citep{rslad}, AdaAD \citep{adaad}, etc.

\section{Experiments}

We explain the experimental setup, followed by comparative performance evaluations of the proposed IGDM method against various AT and AD methods.

\begin{table}[b]
%\begin{adjustbox}{width=\columnwidth,center}
 \centering
\caption{Performance (\%) of the teacher models. Experiments with teacher models in \textit{italics} are provided in the supplementary material.}
 \begin{tabular}{lllccc}
 \toprule
     \multicolumn{1}{c}{Dataset} & \multicolumn{1}{c}{Teacher name} & \multicolumn{1}{c}{Architecture}  & Clean & PGD & AA\\
\midrule

\multicolumn{1}{l}{\multirow{3}{*}{CIFAR-100}} & BDM-AT \citep{wang2023better} & WideResNet-28-10 & 72.58 & 44.24 & 38.83 \\
&  LTD \citep{LTD} & WideResNet-34-10 & 64.07 & 36.61 & 30.57 \\

& \textit{IKL-AT \citep{cui2023decoupled}} & WideResNet-28-10 & 73.80 & 44.14 & 39.18 \\
\midrule

\multicolumn{1}{l}{\multirow{3}{*}{CIFAR-10}} & LTD \citep{LTD} & WideResNet-34-10 & 85.21 & 60.89 & 56.94 \\
 & \textit{BDM-AT \citep{wang2023better}} & WideResNet-28-10  & 92.44 & 70.63 & 67.31 \\
 & \textit{IKL-AT \citep{cui2023decoupled}} & WideResNet-28-10  & 92.16 & 71.09 & 67.73 \\
\midrule
 SVHN & PGD-AT \citep{PGD}& WideResNet-34-10 & 93.90 & 61.78 & 54.28  \\
\midrule
 Tiny-ImageNet & \textit{RiFT \citep{zhu2023improving}} & WideResNet-34-10 & 52.54 & 25.52 & 21.78  \\
 \bottomrule
 \end{tabular}
 % \end{adjustbox}

 \label{tab:main_teacher}

\end{table}

\begin{table*}[h]
 \centering
 \caption{Adversarial distillation result on ResNet-18 with two teacher models on CIFAR-100. The Clean, PGD, FGSM, C\&W, and AA each indicate performance (\%). Bold indicates cases where IGDM improved accuracy by more than 0.5 percentage points.}
 \setlength{\tabcolsep}{5.5pt}
 \begin{tabular}{lcccccccccc}
 \toprule
    \multicolumn{1}{c}{\multirow{2.5}{*}{Method}}  & \multicolumn{5}{c}{ CIFAR-100 with BDM-AT teacher}& \multicolumn{5}{c}{CIFAR-100 with LTD teacher} \\
   \cmidrule(r){2-6}
   \cmidrule(r){7-11}
   & Clean & FGSM & PGD & C\&W & AA & Clean & FGSM & PGD & C\&W & AA \\
\midrule

 PGD-AT  & 55.80 & 23.51 & 19.88 & 20.46 & 18.86 & 55.80 & 23.51 & 19.88 & 20.46 & 18.86  \\
 \textbf{+IGDM} & \textbf{60.83} & \textbf{38.11} & \textbf{35.09} & \textbf{30.21}& \textbf{29.16}& \textbf{60.49} & 37.22 & \textbf{33.72} & \textbf{29.61} & \textbf{28.02} \\
 \midrule
 TRADES & 53.56 & 29.85 & 25.85 & 23.32 & 22.02 & 53.56 & 29.85 & 25.85 & 23.92 & 22.02 \\
 \textbf{+IGDM}  & \textbf{60.88} & \textbf{36.26} & \textbf{32.26} & \textbf{26.50} & \textbf{25.50} & \textbf{59.29} & 36.08 & \textbf{32.08} & \textbf{26.10} & \textbf{26.10} \\
 \midrule
 ARD & 61.51 & 34.23 & 30.23 & 26.97 & 24.77& 61.34 & 35.19 & 31.19 & 27.74 & 25.74\\
 \textbf{+IGDM} & 61.62 & \textbf{39.75} & \textbf{35.75} & \textbf{30.99} & \textbf{28.79} & 61.58 & \textbf{37.45} & \textbf{33.45} & \textbf{29.94} & \textbf{27.84}\\
 \midrule

 IAD & 59.92 & 35.47  & 31.47 & 26.91 & 25.15& 60.12 & 36.91 & 32.91 & 28.29 & 26.29\\
 \textbf{+IGDM} & \textbf{62.99} & \textbf{37.32} & \textbf{34.76} & \textbf{29.55} & \textbf{27.76} & \textbf{62.73} & \textbf{37.69} & \textbf{33.69} & \textbf{29.65} & \textbf{27.49} \\
 \midrule
 AKD & 60.27 & 35.38 & 31.38 & 26.29 & 25.09 & 60.46 & 35.53 & 31.53& 27.29 & 25.37\\
 \textbf{+IGDM} & 60.42 & \textbf{37.63} & \textbf{33.62} & \textbf{29.27} & \textbf{28.86} & 60.31& \textbf{38.15} & \textbf{34.15}& \textbf{29.44} & \textbf{28.44}\\
 \midrule
 RSLAD & 60.22 & 36.16 & 32.16 & 27.96 &26.76 & 60.01 & 36.39 & 32.39 & 28.94 & 26.94 \\
 \textbf{+IGDM} & \textbf{62.06} & \textbf{39.67} & \textbf{35.67} & \textbf{30.98} & \textbf{29.78} & 60.43 & \textbf{38.02} & \textbf{34.02} & \textbf{29.94} & \textbf{28.51} \\
 \midrule
 AdaAD &64.43 & 37.33 & 33.21 & 29.53 & 28.06& 63.34 & 37.39 & 33.39 & 29.73 & 27.81\\
 AdaIAD & 64.13& 37.33 & 33.33& 29.02 & 27.82 & 63.24 & 37.45 & 33.45 & 29.04 & 27.83\\
 \textbf{+IGDM} & 64.44& \textbf{39.31} & \textbf{36.19} & \textbf{31.75} & \textbf{30.32} & 63.44 & \textbf{38.23} & \textbf{34.23} & \textbf{31.09} & \textbf{28.87} \\
\bottomrule
 \end{tabular}
 \label{tab:main_resnet_cifar100}
\end{table*}

\begin{table*}[t]
 \centering
 \caption{Adversarial distillation result on ResNet-18 on CIFAR-10 and SVHN. The Clean, PGD, FGSM, C\&W, and AA each indicate performance (\%). Bold indicates cases where IGDM improved accuracy by more than 0.5 percentage points.}
 \setlength{\tabcolsep}{5.5pt}
 \begin{tabular}{lcccccccccc}
 \toprule
   \multicolumn{1}{c}{\multirow{2.5}{*}{Method}}& \multicolumn{5}{c}{CIFAR-10 with LTD teacher}& \multicolumn{5}{c}{SVHN with PGD-AT teacher} \\
   \cmidrule(r){2-6}
   \cmidrule(r){7-11}
   & Clean & FGSM & PGD & C\&W & AA & Clean & FGSM & PGD & C\&W & AA \\
\midrule PGD-AT  & 84.52 & 52.42& 42.80 & 42.98& 41.12 & 91.62& 65.93& 48.07& 48.34 & 42.46\\
 \textbf{+IGDM} & 84.15 & \textbf{59.21}& \textbf{53.70} & \textbf{51.19}& \textbf{49.63} & \textbf{92.57}& \textbf{72.19}& \textbf{60.55}& \textbf{56.10} & \textbf{52.12}\\
\midrule TRADES  & 82.46 & 56.97& 49.13 & 47.98& 47.09 & 89.91& 69.81& 57.52& 51.32 & 50.74\\
 \textbf{+IGDM} &\textbf{83.50}& \textbf{60.84}& \textbf{54.64} &\textbf{49.34}& \textbf{48.83} &\textbf{91.98}& \textbf{71.85}& \textbf{60.09}& \textbf{56.68} & \textbf{54.15}\\
\midrule
 ARD & 85.04 & 60.31& 53.27 & 50.27& 49.49 & 92.32 & 70.46& 55.62 & 53.07 & 48.24\\
 \textbf{+IGDM} & 85.18 & \textbf{61.24}& \textbf{54.75} & \textbf{51.31}& \textbf{50.20} & 92.19& \textbf{72.01}& \textbf{60.07}& \textbf{56.56} & \textbf{52.95}\\
 \midrule
 IAD & 84.33 & 61.21& 54.24 & 50.97& 50.09 & 91.62& 70.82& 56.42& 53.01 & 47.76\\
 \textbf{+IGDM} & 84.49 & \textbf{62.45}& \textbf{56.55} & \textbf{53.01}& \textbf{51.09} & \textbf{93.03}& \textbf{71.85}& \textbf{58.86}& \textbf{54.66} &\textbf{51.45}\\
 \midrule
 AKD & 85.10& 59.07& 51.53& 49.13& 48.04& 92.49& 70.44& 56.49& 53.91  & 50.17\\
 \textbf{+IGDM} & 84.94& \textbf{61.36}& \textbf{54.89}& \textbf{51.55}& \textbf{50.87}& 92.50& \textbf{71.15}& \textbf{59.38}& \textbf{55.97} & \textbf{52.51}\\
 \midrule
 RSLAD & 83.59 & 60.97& 55.98 & 53.15 & 52.13 & 90.52& 62.74& 53.80& 50.01 & 48.41\\
 \textbf{+IGDM} & 83.67 & \textbf{62.41}& \textbf{57.34} & \textbf{54.00}  & \textbf{53.10} & 90.71& \textbf{64.71}& \textbf{56.14}& \textbf{53.98} & \textbf{50.69}\\
 \midrule
 AdaAD & 84.74& 61.87& 56.78 & 53.51& 52.79 & 93.27 & 67.14& 57.43 & 54.88 & 52.93\\
 AdaIAD & 84.75& 61.98& 57.04& 53.57& 52.88& 93.39& 67.10& 57.26& 54.76 & 52.74\\
 \textbf{+IGDM} & 84.83 & \textbf{62.54} & \textbf{57.61} & \textbf{55.09} & \textbf{54.02} & 93.48 & \textbf{68.34}& \textbf{58.74} & \textbf{56.03} & \textbf{53.89} \\

\bottomrule
 \end{tabular}
 \label{tab:main_resnet_cifar10_svhn}
\end{table*}

\subsection{Settings}
\paragraph{Teacher and Student Models}
We selected three teacher models including  LTD \citep{LTD}, BDM-AT \citep{wang2023better}, and IKL-AT \citep{cui2023decoupled} for CIFAR-10/100.
The LTD model is widely adopted in prior AD research, while the others have achieved high-ranking performance in RobustBench \citep{croce2021robustbench}, demonstrating superior robustness against AutoAttack \citep{AutoAttack}.
For Tiny-ImageNet, we employed a pre-trained RiFT model \citep{zhu2023improving}, and for SVHN, we trained a WideResNet-34-10 \citep{BMVC2016_87} with PGD-AT.
% The detailed information can be found in \Cref{tab:main_teacher}.
For student models, we employed the ResNet-18 \citep{he2016deep} and MobileNetV2 \citep{sandler2018mobilenetv2} for CIFAR-10/100, ResNet-18 for SVHN, and PreActResNet-18 \citep{he2016identity} for Tiny-ImageNet.

\paragraph{Evaluation Metrics}
After training, we evaluate performance using five metrics: Clean, FGSM, PGD, C\&W, and AutoAttack (AA) accuracy. Clean refers to the accuracy on the test dataset. We measure FGSM and PGD accuracy against fast gradient sign method (FGSM) \citep{FGSM} and 20-step projected gradient descent (PGD) attacks \citep{PGD}, respectively. The C\&W attack measures accuracy against \cite{CW_attack}, while AA evaluates accuracy against the AutoAttack method \citep{AutoAttack}. All attacks were conducted within an $l_{\infty}$-norm bound of $8/255$.

\paragraph{Implementation}
We utilized the CIFAR-10/100 \citep{krizhevsky2009learning}, SVHN \citep{svhn}, and Tiny-ImageNet \citep{le2015tiny} datasets for our experiments.
Random crop and random horizontal flip were applied, while other augmentations were not utilized.
Our training methods encompassed conventional adversarial training methods, PGD-AT \citep{PGD} and TRADES \citep{TRADES}, as well as adversarial distillation techniques including ARD \citep{ard}, IAD \citep{iad}, AKD \citep{akd}, RSLAD \citep{rslad}, and AdaAD \citep{adaad}.
In our comparative analysis, we integrated IGDM into all of these methods, and we used shortened notations to represent the methods with IGDM. 
For example, when IGDM is combined with ARD, we denote it as ARD \textbf{+ IGDM} or IGDM$_{\text{ARD}}$.
We employed the recommended inner loss functions for generating adversarial examples as outlined in each baseline as specified in \Cref{sec:apendix_training_detail}.
We fixed the hyperparameters of all given methods used in the experiment as original paper.
Then, we adjusted the hyperparameter $\alpha$ of IGDM.
For the surrogate loss of IGDM, we employed KL divergence loss, but using alternative surrogate losses, such as L1 and L2 loss, yields nearly indistinguishable results.
The detailed experimental setting can be found in \Cref{sec:apendix_training_detail}.

\subsection{Results}
\paragraph{Main Results}
We present the comprehensive results of integrating IGDM into other baseline methods and their original versions in \Cref{tab:main_resnet_cifar100}, \Cref{tab:main_resnet_cifar10_svhn}.
More experimental results are in the appendix: distillation results on MobileNetV2 student (\Cref{sec:mobilenet}), distillation with IKL-AT teacher (\Cref{sec:IKL-AT}), and experiments on the Tiny-ImagNet dataset (\Cref{sec:diff_dataset}).
IGDM significantly improves robustness against various attacks with consistent clean accuracy, regardless of the original methods, datasets, student models, or teacher models.
IGDM demonstrates notably enhanced AA robustness on the CIFAR-100 dataset, achieving 30.32\% on ResNet-18 with a BDM-AT teacher.
%and 29.32\% on MobileNetV2.

% Furthermore, on the Tiny-ImageNet dataset, IGDM achieves AA robustness of 21.17\%, which is nearly equivalent to that of the teacher model.

\begin{figure}[h]
\centering
    \includegraphics[width=0.95\textwidth] {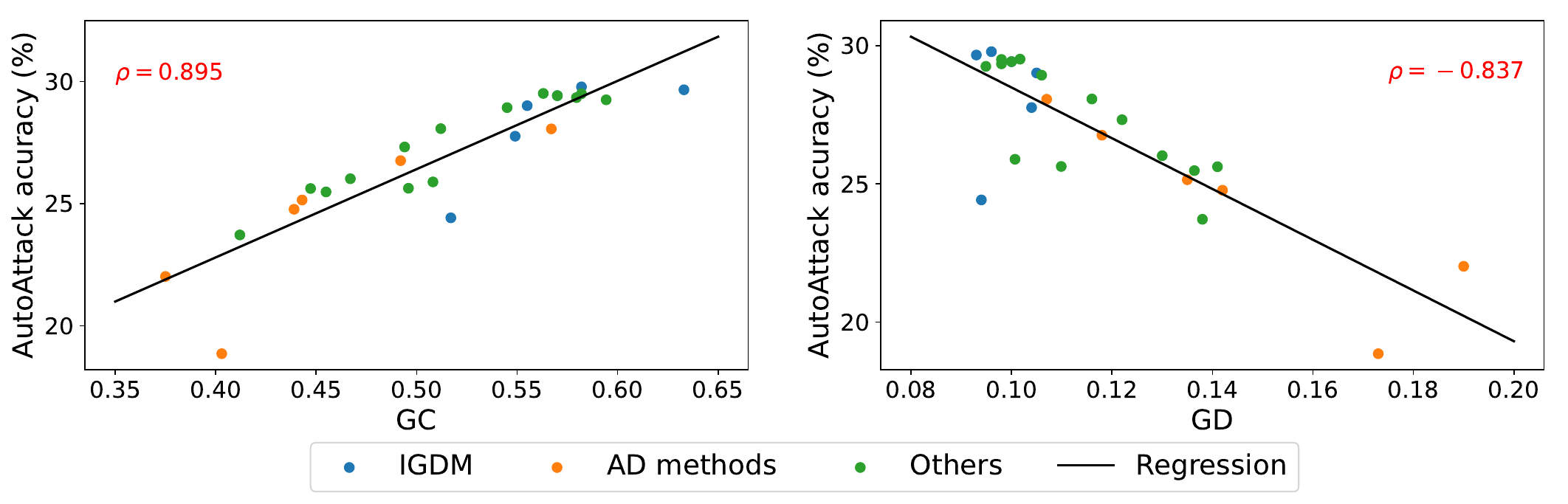}%
    \vspace{-0.2cm}
    \caption{Correlation between GC and AA ($left$) and between GD and AA ($right$). All results were obtained using ResNet-18 and a BDM-AT teacher on CIFAR-100. The values for IGDM and AD methods match those in \Cref{tab:main_resnet_cifar100}, while 'Others' represent results from additional experiments under the same configuration. $\rho$ denotes the correlation coefficient.}
    \label{fig:regression}
    \vspace{-0.5cm}
\end{figure}

\paragraph{Gradient Alignment}
\begin{wraptable}{r}{7.5cm}
\vspace{-0.5cm}
\caption{Gradient alignment on CIFAR-100, using the ResNet-18 student model. 
The numbers in bold indicate enhanced gradient alignment.}
 \centering
 \setlength{\tabcolsep}{3pt}
 \begin{tabular}{lcccccc}
 \toprule
     \multicolumn{1}{c}{\multirow{2.5}{*}{Method}} &  \multicolumn{3}{c}{BDM-AT}&\multicolumn{3}{c}{LTD}\\

     \cmidrule(r){2-4} \cmidrule(r){5-7}  &AA& GD$\Downarrow$& GC$\Uparrow$ &AA& GD$\Downarrow$& GC$\Uparrow$\\ 
\midrule
 ARD  &24.77 &0.142&0.439&25.74 & 0.108& 0.592\\
\textbf{+IGDM}  &\textbf{28.79} &\textbf{0.101}&\textbf{0.571}&\textbf{27.84} & \textbf{0.082}& \textbf{0.688}\\
 \midrule
 IAD  &25.15 &0.135&0.443&26.29 & 0.102& 0.596\\
\textbf{+IGDM}   &\textbf{27.76} &\textbf{0.104}&\textbf{0.549}&\textbf{27.49} & \textbf{0.086}& \textbf{0.674}\\
 \midrule
 AKD &25.09 & 0.127 & 0.438 &25.37& 0.113& 0.584\\
 \textbf{+IGDM}   &\textbf{28.86}& \textbf{0.114} & \textbf{0.513}&\textbf{28.44}& \textbf{0.078}& \textbf{0.693}\\
 \midrule
 RSLAD  &26.76 &0.118&0.492&26.94 & 0.089& 0.658\\
\textbf{+IGDM}  &\textbf{29.78} &\textbf{0.096}&\textbf{0.582}&\textbf{28.51} & \textbf{0.077} & \textbf{0.709} \\
  \midrule
 AdaAD  &28.06 &0.107&0.567&27.81 & 0.077& 0.736\\
 AdaIAD &27.82& 0.107& 0.568&27.83 & 0.077 &0.733\\
\textbf{+IGDM}  &\textbf{30.32} &\textbf{0.086} &\textbf{0.643} &\textbf{28.87} &  \textbf{0.070} & \textbf{0.769} \\
 \bottomrule
 \end{tabular}
 \label{tab:gd_gc}
 \vspace{-0.5cm}
\end{wraptable}
We justify that IGDM can align the student's input gradient with the teacher's input gradient.
To quantify this alignment between the two input gradients, we introduce two metrics: mean Gradient Distance (GD) and mean Gradient Cosine similarity (GC). 
GD quantifies the average L2 distance between the input gradients of the teacher and student models for all test samples, with smaller values indicating closer alignment. GC measures the cosine similarity between these gradients, where a value closer to one signifies better alignment.
In \Cref{tab:gd_gc}, IGDM improves gradient alignment with the teacher model, with improved robustness. 
As shown in \Cref{fig:regression}, the robustness of the student model increases as GD decreases and GC rises, confirming the positive correlation between gradient alignment and student model performance.

\paragraph{Point-wise Alignment}
% Input gradient of adversarially trained models has semantically meaningful information, aligning with human perception \citep{tsipras2018robustness, engstrom2019adversarial, srinivas2023models}.
% Thus, it is natural to argue that distilling input gradients offers additional information for the student to effectively mimic the robust teacher.
Existing adversarial distillation methods have largely overlooked the importance of input gradients, instead focusing on point-wise alignment.
For example, RSLAD \citep{rslad} and AdaAD \citep{adaad} explicitly aim to align clean and adversarial outputs with the teacher's clean output.
Thus, we reinterpret point-wise alignment in terms of input gradient matching, where IGDM demonstrates superior performance compared to existing AD methods in achieving this alignment.
 % IGDM implicitly aligns the student and teacher by matching their input gradients. This highlights the broader perspective of gradient matching as a key component of alignment in adversarially trained models.
% We elucidate the role of gradient matching in facilitating students to emulate their teachers.
% An underlying aim of knowledge distillation is to foster alignment between the teacher and the student.
% In AD, the alignment should consider not only clean images but also adversarial ones.
% For instance, RSLAD \citep{rslad} attempts to match both clean and adversarial outputs to the clean output of the teacher model, and
% AdaAD \citep{adaad} points out there is a non-negligible property of local variance on the teacher model, aligning the student and teacher under the local variance.
% While RSLAD and AdaAD explicitly try to match the output between teacher and student, IGDM implicitly matches teacher and student by matching the input gradient.
We define point-wise distance $D(\mathbf{x},\boldsymbol{\delta}) = \lVert f_{T}(\mathbf{x} + \boldsymbol{\delta}) - f_S(\mathbf{x} + \boldsymbol{\delta}) \rVert $.
Then given $\mathbf{x}$, the upper bound of $D(\mathbf{x},\boldsymbol{\delta})$ for sufficiently small $\boldsymbol{\delta}$ is as follows.
\begin{equation}\label{dpoint}
\begin{split}
    D(\mathbf{x},\boldsymbol{\delta}) \leq \left \lVert(f_{T}(\mathbf{x}) - f_{S}(\mathbf{x}) \right \rVert  + \left \lVert \left (\frac{\partial f_{T}(\textbf{x})}{\partial \textbf{x}} - \frac{\partial f_{S}(\textbf{x})}{\partial \textbf{x}}\right )^T \boldsymbol{\delta} \right \rVert. \\  
\end{split}
\end{equation}
% We see that the distance $D(\mathbf{x},\boldsymbol{\delta})$ is bounded by the output term and the gradient term.
Hence, the better the gradient matching, the smaller the upper bound of $D(\mathbf{x},\boldsymbol{\delta})$ becomes, aligning the teacher and student in a point-wise manner.

% We argue in \Cref{dpoint} that IGDM facilitates point-wise alignment.
To empirically substantiate this assertion, we compared the distances between the teacher and the student models.
We calculated $D(f_S(\mathbf{x} + \boldsymbol{\delta}), f_T(\mathbf{x} + \boldsymbol{\delta})) $ with L2 distance for random noise from the uniform distribution and adversarial perturbation in \Cref{tab:ablation_dpoint}.
For the adversarial noise, we conducted an adversarial attack using the inner maximization loss corresponding to each method, with a fixed number of steps set to 20.
In all cases, we observed an improvement in alignment when using IGDM.
In particular, IGDM enhanced alignment against not only adversarial noise but also random noise.
This outcome demonstrates that our module contributes to point-wise alignment, improving overall robustness by significantly reducing local invariance during training, as asserted in \cite{adaad}.

% \begin{figure}[h]
%     \centering
%     \begin{subfigure}[b]{0.45\textwidth}
%         \includegraphics[width=\textwidth]{figure/point_wise.pdf}
%         \caption{Point-wise distance with various $\epsilon$.}
%         \label{fig:point_wise}
%     \end{subfigure}
%     \hfill
%     \begin{subfigure}[b]{0.45\textwidth}
%         \includegraphics[width=\textwidth]{figure/various_eps_pgd.pdf}
%         \caption{PGD accuracy with various $\epsilon$.}
%         \label{fig:various_eps_pgd}
%     \end{subfigure}
%     \caption{}
%     \label{fig:overall}
% \end{figure}

\begin{table}[h]
\caption{Point-wise alignment with IGDM on CIFAR-100, using the ResNet-18 student model and two teacher models. The $\boldsymbol{\delta} \sim U$ indicates uniformly selected $\boldsymbol{\delta}$ from $U[-8/255,8/255]$. 
The numbers in bold indicate enhanced point alignment.}
 \centering
 \setlength{\tabcolsep}{13pt}
 \begin{tabular}{lccccll}
 \toprule
     \multicolumn{1}{c}{\multirow{2.5}{*}{Method}}  &\multicolumn{2}{c}{BDM-AT Teacher}&\multicolumn{2}{c}{LTD teacher}&\multicolumn{2}{c}{IKL-AT teacher}\\

     \cmidrule(r){2-3} \cmidrule(r){4-5} \cmidrule(r){6-7}& $\boldsymbol{\delta} \sim U$ & $\boldsymbol{\delta} = $ adv   & $\boldsymbol{\delta} \sim U$ & $\boldsymbol{\delta} = $ adv  & $\boldsymbol{\delta} \sim U$ &$\boldsymbol{\delta} = $ adv  \\ 
\midrule
 ARD &0.2392 & 0.2848 & 0.1822 & 0.2176  & 0.4030 &0.4734\\
\textbf{+IGDM} &\textbf{0.1312} & \textbf{0.1533} & \textbf{0.1301} & \textbf{0.1501}  & \textbf{0.1504} &\textbf{0.1690} \\
 \midrule
 IAD &0.2374 &0.2811 & 0.1905 & 0.2190  & 0.3939 &0.4568 \\
\textbf{+IGDM}  &\textbf{0.1342} &\textbf{0.1684} & \textbf{0.1194} & \textbf{0.1395}  & \textbf{0.1524} &\textbf{0.1971} \\
 \midrule
% 0.3196 & 0.3184 & 0.3622
% 0.1716 & 0.1704 & 0.2081
 AKD& 0.3184 & 0.3622 &  0.2017 & 0.2323 & 0.4325 &0.4810 \\
 \textbf{+IGDM}  & \textbf{0.1704} & \textbf{0.2081} & \textbf{0.1156}  & \textbf{0.1312} & \textbf{0.1521}& \textbf{0.1698}\\
 \midrule
 RSLAD &0.1903 &0.2375 & 0.1686 & 0.2019  & 0.2227 &0.2856 \\
\textbf{+IGDM} &\textbf{0.1233} &\textbf{0.1516} & \textbf{0.1219} & \textbf{0.1424}  & \textbf{0.1361} &\textbf{0.1681} \\
  \midrule
 AdaAD &0.1794 &0.2529 & 0.1348 & 0.2028  & 0.2499 &0.3400 \\
  AdaIAD &0.1803 &0.2518 & 0.1331 & 0.2010  & 0.2504 &0.3479 \\
\textbf{+IGDM} &\textbf{0.1294} &\textbf{0.1857} & \textbf{0.1253} & \textbf{0.1833}  & \textbf{0.1338} &\textbf{0.2002} \\
 \bottomrule
 \end{tabular}
 \label{tab:ablation_dpoint}
\end{table}

\subsection{Comparison with the State-of-the-Art Method: AdaAD}

\paragraph{Simple Inner Maximization with IGDM} AdaAD \citep{adaad} employs a teacher model for inner maximization to enhance student model robustness. 
However, this approach significantly increases computational overhead due to the teacher model's large architecture. 
In contrast, IGDM achieves competitive robustness without a teacher model in inner maximization, simplifying the overall training process and reducing training time.
In \Cref{tab:AdaAD_time}, we compare performance with and without the teacher model for inner maximization. 
When AdaAD's inner maximization is replaced with a PGD attack on the student model alone, robustness drops significantly. However, IGDM$_{\text{AdaAD}}$ achieves superior robustness more than the original AdaAD method, with reduced training time by a factor of three.
This highlights the efficiency of IGDM, offering both robustness and flexibility by eliminating the need for a teacher model on inner maximization.

\begin{table}[h]
 \centering
 \caption{AutoAttack accuracy (\%) and computational overhead of AdaAD and IGDM$_{\text{AdaAD}}$ with different inner loss on CIFAR-100 with the ResNet-18 student. `T/E' and `Mem' refer to time per epoch (in minutes) and memory usage.
 The `w/o $T_{in}$' indicates that the inner loss is computed using a PGD attack solely on the student model, without the teacher's involvement.
}
 \setlength{\tabcolsep}{5pt}
 \begin{tabular}{lrrrrrrrrr}
 \toprule
 \multicolumn{1}{c}{\multirow{2}{*}{Method}}  & \multicolumn{3}{c}{BDM-AT teacher}&  \multicolumn{3}{c}{LTD teacher} & \multicolumn{3}{c}{IKL-AT teacher}\\
\cmidrule(r){2-4}\cmidrule(r){5-7}\cmidrule(r){8-10} & AA & T/E$\Downarrow$  &Mem$\Downarrow$& AA & T/E$\Downarrow$ &Mem$\Downarrow$ & AA & T/E$\Downarrow$ &Mem$\Downarrow$ \\
 \midrule
AdaAD  & 28.06 & 10.62 
&4711M& 27.81 &12.12 
& 4900M& 26.89& 10.91&4711M
\\
AdaAD w/o $T_{in}$ &  
25.62 & 2.26 
&2063M&  
26.57 &2.42 
& 2634M& 23.37& 2.27&2063M
\\
IGDM$_{\text{AdaAD}}$ &  30.32 & 11.21 
&4711M&  28.87 &13.27 
& 4900M & 29.22& 11.30&4711M
\\
IGDM$_{\text{AdaAD}}$ w/o $T_{in}$ &  
 29.34 & 3.36  &2611M&  
 28.12 &3.64  & 3187M& 28.94& 3.38&2611M\\
 \bottomrule
 \end{tabular}

 \label{tab:AdaAD_time}
 % \vspace{-0.1cm}
\end{table}

\paragraph{Comparison of IGDM and AdaAD under Same Experimental Conditions} We replicated AdaAD's experimental setup using the same teacher-student model configuration for a fair comparison. Specifically, we used the CIFAR-100 dataset with the LTD teacher and the CIFAR-10 dataset with the LTD$_2$ teacher. (In the AdaAD paper, the LTD teacher was trained on WideResNet-34-20 for CIFAR-10, so we refer to it as LTD$_2$ to reflect this difference.)
Both setups were evaluated with a ResNet-18 student, as shown in \Cref{tab:adaad_paper}. 
While the AdaAD paper proposed two methods, AdaAD and AdaIAD, and noted a performance gap of about 1 percentage point, our experiments showed a smaller difference.
As a result, we focused on applying the module specifically to AdaAD in our implementation.
Notably, IGDM$_{\text{AdaAD}}$ consistently demonstrated a significant improvement in robustness accuracy. 
This demonstrates that even in a fair comparison, the IGDM module consistently enhances adversarial robustness, underscoring its effectiveness.

\begin{table}[h]
 \centering
 \caption{Comparison between AdaAD paper results and our implementation results with identical experimental settings.}
 \setlength{\tabcolsep}{7.5pt}
 \begin{tabular}{lccrcrc}
 \toprule
 \multicolumn{1}{c}{\multirow{2}{*}{Method}} & \multicolumn{2}{c}{Result from} & \multicolumn{2}{c}{CIFAR-100 with LTD}& \multicolumn{2}{c}{CIFAR-10 with LTD$_\text{2}$}\\
 \cmidrule(r){2-3}\cmidrule(r){4-5}\cmidrule(r){6-7}
  & \cite{adaad}& Ours & Clean & AA  & Clean &AA  \\

 \midrule
 
 AdaAD & \checkmark & & 62.19 & 26.74 & 85.58&51.37\\
 AdaIAD & \checkmark & & 62.49 & 27.98 & 85.04&52.96\\
 AdaAD &  & \checkmark & 63.34 & 27.81 & 85.47&52.47\\
  AdaIAD& & \checkmark & 63.24&27.83 & 85.34&52.30\\
 IGDM$_{\text{AdaAD}}$ &  & \checkmark & 63.44& \textbf{28.87} & 85.50& \textbf{53.45}\\
\bottomrule
 \end{tabular}

 \label{tab:adaad_paper}
  % \vspace*{-0.3cm}
\end{table}

% \begin{wrapfigure}{r}{0.53\textwidth}
%     \centering
%     \begin{minipage}[c]{.23\textwidth}
%       \centering
%       \includegraphics[scale=0.2]{figure/compared_adaad_fixed.pdf}
%     \end{minipage}%
%     \begin{minipage}[c]{.3\textwidth}
%       \centering
%       \includegraphics[scale=0.2]{figure/compared_adaad_grad_fixed.pdf}    
%     \end{minipage}
%       \\
%     \vspace{-0.2cm}
%     \begin{minipage}[t]{.25\textwidth}
%       \subcaption{}   
%        \label{fig:ratio_teacher}
%     \end{minipage}%
%     \begin{minipage}[t]{.25\textwidth}
%         \subcaption{}
%        \label{fig:ratio_student}%
%     \end{minipage}
%     \caption{Performance over hyperparameter adjustment matching the clean output of AdaAD and the gradient distance from the teacher model.}%
%     \label{fig:adaad_hyper_figure}%
%     \vspace{-0.5cm}
% \end{wrapfigure}

\begin{wraptable}{r}{0.5\textwidth}
 \centering
 \vspace{-0.4cm}
 \caption{Comparison between AdaAD with various hyperparameter.}
    \setlength{\tabcolsep}{3pt}
 \begin{tabular}{lcccc}
  \toprule

Method & Clean &  AA & GD$\Downarrow$ & GC$\Uparrow$ \\
\midrule
AdaAD($\lambda=0.25)$& 66.48 & 25.51 & 0.118  &  0.526 \\
AdaAD($\lambda=0.50)$& 65.44 & 27.24 & 0.108  &  0.564 \\
AdaAD($\lambda=0.75)$& 64.98 & 27.42 & 0.109  &  0.560 \\
AdaAD($\lambda=1.0)$& 64.43 & 27.82 & 0.107  & 0.568 \\
\midrule
 IGDM$_{\text{AdaAD}}$& 64.44 & \textbf{30.32} & \textbf{0.086}  & \textbf{0.643} \\
 \bottomrule
 \end{tabular}
 \label{tab:hyper_adaad}
\end{wraptable}
\paragraph{Limitations of Hyperparameter Tuning in AdaAD for Gradient Alignment} AdaAD utilizes a single hyperparameter, $\lambda$, to control the balance between the distillation of adversarial and clean inputs, as $\lambda \cdot \text{KL}(f_S(\textbf{x} + \boldsymbol{\delta}) \| f_T(\textbf{x} + \boldsymbol{\delta})) + (1 - \lambda) \cdot \text{KL}(f_S(\textbf{x}) \| f_T(\textbf{x}))$. However, in their implementation, $\lambda$ is always set to one.
Initially, we hypothesized that adjusting $\lambda$ could facilitate the distillation of both adversarial and clean inputs, potentially aligning the gradients.
Yet, as shown in \Cref{tab:hyper_adaad}, modifying $\lambda$ degraded both robustness and gradient matching.
This suggests that while point-wise matching is achieved in AdaAD, simply distilling two points is insufficient for capturing the teacher’s gradients. 
In contrast, our approach enables effective gradient alignment and better distillation of the teacher model’s robustness.

% \begin{table}[h]
%  \centering
%  \caption{The robust accuracy of unseen norm attacks and a black-box attack. The numbers in bold indicate improved performance.}
%    \renewcommand{\arraystretch}{1.03}
%     \setlength{\tabcolsep}{5pt}
%  \begin{tabular}{lcc}
%   \toprule
% \multicolumn{1}{c}{\multirow{2}{*}{Method}} & \multicolumn{2}{c}{Unseen}  \\ 
% \cmidrule(r){2-3} 
% \cmidrule(r){4-4} 
% & \multicolumn{1}{l}{PGD-L1} & \multicolumn{1}{l}{PGD-L2} 

% \midrule

%  0.25 & 43.11 & 39.51 \\
% 0.5 & 43.11 & 39.51 \\
% 0.75 & 43.11 & 39.51 \\
% 1.0 & 43.11 & 39.51 \\
%  \bottomrule
%  \end{tabular}
%  \label{tab:unseen attack}
% \end{table}

% AdaAD also proposed that increasing the $\epsilon$ boundary for inner maximization enhances robustness against AutoAttack. In their CIFAR-10 experiment with a ResNet-18 student, they reported the highest robustness of 54.23\% for $\epsilon = 32/255$. Applying the same setting to IGDM$_\text{AdaAD}$, we achieved a higher AutoAttack accuracy of 54.74\%, with clean accuracy of 83.76\% and PGD accuracy of 58.69\%. This demonstrates that the IGDM module improves robustness and surpasses AdaAD under identical conditions.

\subsection{Ablation Studies}
In this section, we conduct more extensive experiments, including robustness against unseen attacks and distillation results with various teacher models.
Here, we chose the CIFAR-100 dataset with ResNet-18 architecture of student model.
% In the case of the teacher, we selected BDM-AT.

\paragraph{Robustness against Unseen Attacks}
\begin{wraptable}{r}{0.45\textwidth}
 \centering
 \vspace{-0.4cm}
 \caption{The CIFAR-100 performance, evaluated with OODRobustBench.}
    \setlength{\tabcolsep}{3pt}
 \begin{tabular}{lccc}
  \toprule
\multicolumn{1}{c}{\multirow{2}{*}{Method}} &Clean Acc (\%)&\multicolumn{2}{c}{Robust Acc (\%)}\\ 
\cmidrule(r){2-2} 
\cmidrule(r){3-4} 
& \multicolumn{1}{c}{$\text{OOD}_d$} & $\text{OOD}_d$ & $\text{OOD}_t$ \\
\midrule
%   PGD-AT & 44.58 & 11.46  &  10.04 \\
% + \textbf{IGDM} & 49.27 & 19.21  & 17.82 \\ 
% \midrule
%  TRADES  & 43.86 & 14.92  & 12.45 \\
% + \textbf{IGDM} & 48.91 & 15.73 & 17.32 \\ 
% \midrule
 ARD & 49.92 & 15.70 & 14.91 \\
 + \textbf{IGDM} & \textbf{50.65} & \textbf{16.98} & \textbf{17.36} \\
 \midrule
 IAD & 48.75 & 16.21 & 15.03 \\
 + \textbf{IGDM} & \textbf{51.43} & \textbf{18.20} & \textbf{17.35} \\
  \midrule
 AKD & 49.74 & 15.95 & 14.96 \\
 + \textbf{IGDM} & 49.83 & \textbf{18.52} & \textbf{17.07} \\
   \midrule
  RSLAD & 48.96 & 17.41 & 16.46 \\
 + \textbf{IGDM} & \textbf{50.45} & \textbf{19.66} & \textbf{18.53} \\
   \midrule
   AdaAD & 51.53 & 17.77 & 17.32 \\
  AdaIAD & 51.18  & 17.75 & 17.26 \\
   +  \textbf{IGDM} & 51.52  & \textbf{19.66} & \textbf{18.65} \\
 \bottomrule
 \end{tabular}
 \label{tab:unseen attack}
\end{wraptable}
In \Cref{tab:unseen attack}, we measured performance against the different attack scenarios in out-of-distribution (OOD) to demonstrate that we effectively distill the robustness of a teacher model.
OODRobustBench \citep{li2023oodrobustbench} is designed to simulate real-world distribution shifts and evaluate adversarial robustness. It focuses on two types of shifts: dataset shifts ($\text{OOD}_{d}$) and threat shifts ($\text{OOD}_{t}$), offering a more comprehensive assessment compared to relying solely on AutoAttack accuracy.
The term $\text{OOD}_{d}$ encompasses natural and corruption shifts, which consist of variant datasets of CIFAR-100 and common corruptions such as noise and blur.
Meanwhile, $\text{OOD}_{t}$ considers six unforeseen attacks, such as the Recolor \citep{recolor} and StAdv \citep{stAdv_2018}, etc.
Overall, our module improves the performance of existing adversarial distillation methods against various noise and unseen attacks.

\paragraph{Adversarial Distillation with Various Teacher Model}
\begin{wrapfigure}{r}{0.49\textwidth}
\centering
    \vspace{-0.3cm}
     \includegraphics[width=0.5\textwidth] {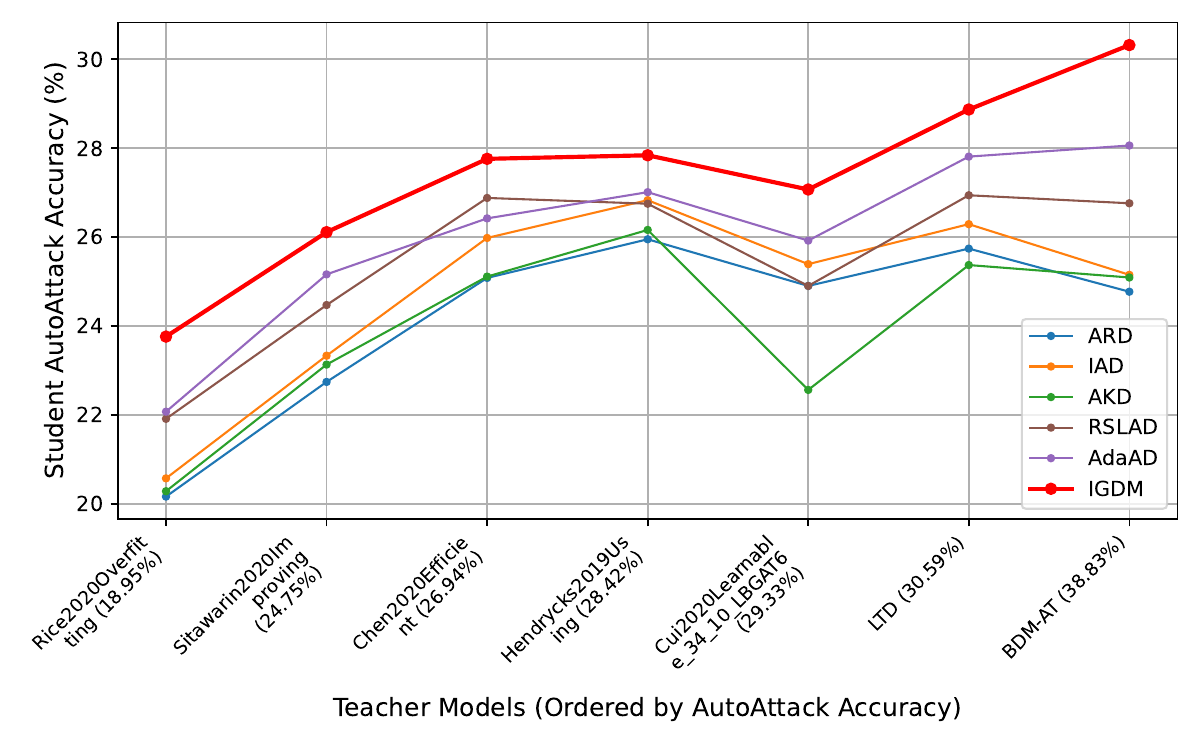}%
     \vspace{-0.4cm}
     \caption{Performance comparison of different adversarial distillation methods across various teacher models.}
     \label{fig:diff_teacher}
     %\vspace{-0.3cm}
 \end{wrapfigure}
We conducted adversarial distillation using several teacher models to further validate the effectiveness of our approach.
We utilized multiple teacher models that were adversarially trained on the CIFAR-100 dataset. 
All these models are publicly available in RobustBench \citep{croce2021robustbench}, and we selected them based on varying levels of robustness against AutoAttack \citep{AutoAttack}.
As shown in the \Cref{fig:diff_teacher}, IGDM consistently outperforms other distillation methods across different teacher setups. 
Notably, even when using less robust teachers, there is a substantial performance gap.
In contrast, while AdaAD demonstrates significant improvements with strong teachers compared to other baselines, it shows minimal differences when employing weaker teachers. 
This highlights that our approach maintains meaningful performance advantages across all teacher models.

\section{Conclusion}

We have proposed a novel method Indirect Gradient Distillation Module (IGDM) for adversarial distillation.
In contrast to conventional adversarial distillation methods that primarily focus on distilling the logits of the teacher model, we concentrate on distilling the gradient information of the teacher model.
We obtain these gradients indirectly by leveraging the locally linear property, a characteristic of adversarially trained models.
Notably, IGDM can be seamlessly applied to existing adversarial distillation methods.
Extensive experimental results demonstrate that the student model with IGDM successfully follows the gradients of the teacher model, resulting in significantly enhanced robustness.

\bibliography{main}
\bibliographystyle{iclr2025_conference}

\newpage
\appendix
\section{Further Experiments}
We conduct further experiments to corroborate our main contribution.
These include distillation results on MobileNetV2 student (\Cref{sec:mobilenet}), distillation with IKL-AT teacher (\Cref{sec:IKL-AT}), and experiments on the Tiny-ImagNet dataset (\Cref{sec:diff_dataset}). We also analyze the drawbacks of direct gradient matching (\Cref{sec:direct_matching}) and the role of logit difference (\Cref{sec:logit_diff}).  Finally, hyperparameter tuning experiments are detailed in \Cref{sec:hyperparameter}.
% , and more detailed experiments on outer minimization, inner maximization, and point-wise alignment of IGDM in \Cref{sec:outer_min}, \Cref{sec:inner_max}, and \Cref{sec:point-wise}.

\subsection{Adversarial Distillation on MobileNetV2 Architecture}
\label{sec:mobilenet}
\begin{table*}[h]
 \centering
 \caption{Adversarial distillation results on MobileNetV2 with BDM-AT and LTD teacher models on CIFAR-100. Bold indicates cases where IGDM improved accuracy by more than 0.5 percentage points or better gradient matching.}
 \setlength{\tabcolsep}{6pt}
 \begin{tabular}{lcccccccccc}
 \toprule
    \multicolumn{1}{c}{\multirow{2.5}{*}{Method}}  & \multicolumn{5}{c}{ CIFAR-100 with BDM-AT teacher}& \multicolumn{5}{c}{CIFAR-100 with LTD teacher} \\
   \cmidrule(r){2-6}
   \cmidrule(r){7-11}
   & Clean & PGD & AA & GD$\Downarrow$ & GC$\Uparrow$ & Clean & PGD & AA & GD$\Downarrow$ & GC$\Uparrow$ \\
\midrule

 PGD-AT  & 59.23 & 24.04 & 21.58 & 0.204 & 0.408 
& 59.23 & 24.04 & 21.58 & 0.194 & 0.486 
\\
 \textbf{+IGDM} & 59.52 & \textbf{32.88} & \textbf{27.28} & \textbf{0.109} & \textbf{0.528} 
& 59.70 & \textbf{32.59} & \textbf{27.14} & \textbf{0.091} & \textbf{0.636} 
\\
 \midrule
 TRADES & 51.05 & 24.83 & 20.62 & 0.132 & 0.377 
& 51.05 & 24.83 & 20.62 & 0.119 & 0.460 
\\
 \textbf{+IGDM}  & \textbf{57.80} & \textbf{29.47} & \textbf{22.05} & \textbf{0.097} & \textbf{0.469} 
& \textbf{56.81} & \textbf{29.87} & \textbf{23.38} & \textbf{0.081} & \textbf{0.640} 
\\
 \midrule
 ARD & 60.74 & 29.92 & 24.33 & 0.137 & 0.445  
& 60.55 & 30.82 & 25.28 & 0.111 & 0.569 
\\
 \textbf{+IGDM} & 60.83 & \textbf{33.55} & \textbf{27.50} & \textbf{0.107} & \textbf{0.534}  
& 60.44 & \textbf{33.36} & \textbf{27.59} & \textbf{0.082} & \textbf{0.681} 
\\
 \midrule

 IAD & 56.35 & 28.96 & 23.43 & 0.129 & 0.430
& 56.11 & 29.55 & 24.22 & 0.106 & 0.547 
\\
 \textbf{+IGDM} & \textbf{57.65} & \textbf{31.97} & \textbf{25.60} & \textbf{0.106} & \textbf{0.506} 
& \textbf{58.60} & \textbf{31.36} & \textbf{25.64} & \textbf{0.090} & \textbf{0.617} 
\\
 \midrule
 AKD & 60.84& 29.37& 24.22& 0.135& 0.428& 60.65& 29.84& 24.85& 0.112& 0.564\\
 \textbf{+IGDM} & 60.39& \textbf{33.62}& \textbf{28.04}& \textbf{0.104}& \textbf{0.538}& 59.94& \textbf{33.70}& \textbf{28.12}& \textbf{0.077}& \textbf{0.715}\\
 \midrule
 RSLAD & 61.29 & 31.74 & 26.18 & 0.121 &0.490  
& 60.43 & 32.37 & 26.85 & 0.092 & 0.632 
\\
 \textbf{+IGDM} & 61.48 & \textbf{35.22} & \textbf{29.32} & \textbf{0.101} & \textbf{0.560}  
& 
60.36 & \textbf{33.99} & \textbf{28.00} & \textbf{0.079} & \textbf{0.696} 
\\
 \midrule
 AdaAD &61.89 & 29.54 & 23.80 & 0.118 & 0.497  
& 61.43& 30.58 & 25.03 & 0.091 & 0.649 
\\
 AdaIAD & 60.83& 29.94& 23.83& 0.116& 0.499& 
61.22& 30.78& 25.22& 0.088& 0.652\\
 \textbf{+IGDM} & 61.43 & \textbf{33.51} & \textbf{27.43} & \textbf{0.097} & \textbf{0.583} & 61.57& \textbf{33.14} & \textbf{27.65} & \textbf{0.076} & \textbf{0.721} \\
\bottomrule
 \end{tabular}
 \label{tab:MNV2_cifar100}
\end{table*}

\begin{table*}[h]
 \centering
 \caption{Adversarial distillation results on MobileNetV2 with BDM-AT and LTD teacher models on CIFAR-10. Bold indicates cases where IGDM improved accuracy by more than 0.5 percentage points or better gradient matching.}
 \setlength{\tabcolsep}{6pt}
 \begin{tabular}{lcccccccccc}
 \toprule
   \multicolumn{1}{c}{\multirow{2.5}{*}{Method}}& \multicolumn{5}{c}{CIFAR-10 with BDM-AT teacher}& \multicolumn{5}{c}{CIFAR-10 with LTD teacher} \\
   \cmidrule(r){2-6}
   \cmidrule(r){7-11}
   & Clean & PGD & AA & GD$\Downarrow$ & GC$\Uparrow$ & Clean & PGD & AA & GD$\Downarrow$ & GC$\Uparrow$ \\
\midrule PGD-AT  & 83.52 & 44.47 & 41.19 & 0.175 & 0.406  
& 83.52 & 44.47 & 41.19 & 0.171 & 0.505 
\\
 \textbf{+IGDM} & 82.78 & \textbf{51.54} & \textbf{47.13} & \textbf{0.074} & \textbf{0.501} 
& 81.73 & \textbf{52.63} & \textbf{48.51} & \textbf{0.048} & \textbf{0.671} 
\\
\midrule TRADES  & 81.57 & 50.49 & 46.88 & 0.074 & 0.492 
& 81.57 & 50.49 & 46.88 & 0.062 & 0.605 
\\
 \textbf{+IGDM} &\textbf{82.33} & \textbf{52.94} & \textbf{47.70} &\textbf{0.050} & \textbf{0.502} 
&81.82 & \textbf{52.96} & \textbf{47.47} & \textbf{0.039} & \textbf{0.645} 
\\
\midrule
 ARD & 84.38 & 48.26 & 44.02 & 0.091 & 0.371  
& 84.18 & 52.16 & 48.11 & 0.054 & 0.578 
\\
 \textbf{+IGDM} & 84.53 & \textbf{52.45} & \textbf{45.06} & \textbf{0.071} & \textbf{0.467}  
& 84.25 & \textbf{54.03} & \textbf{49.53} & \textbf{0.042} & \textbf{0.684} 
\\
 \midrule
 IAD & 83.79 & 48.36 & 44.02 & 0.088 & 0.385  
& 83.38 & 52.71 & 48.45 & 0.051 & 0.592 
\\
 \textbf{+IGDM} & 84.27 & \textbf{52.42} & \textbf{45.08} & \textbf{0.063} & \textbf{0.497}  
& 83.65 & \textbf{54.57} & \textbf{50.14} & \textbf{0.038} &\textbf{0.704} \\
 \midrule
 AKD & 83.75& 47.15& 43.43& 0.093& 0.368& 84.46& 50.84& 46.96& 0.057& 0.566\\
 \textbf{+IGDM} & 83.29& \textbf{53.29}& \textbf{48.41}& \textbf{0.064}& \textbf{0.527}& 83.98& \textbf{54.40}& \textbf{49.99}& \textbf{0.040}& \textbf{0.691}\\
 \midrule
 RSLAD & 84.69 & 50.94 & 47.38 & 0.074 & 0.491 
& 82.60 & 54.55 & 50.44 & 0.040 & 0.706 
\\
 \textbf{+IGDM} & 84.64 & \textbf{52.33} & \textbf{49.27} & \textbf{0.068} & \textbf{0.515} 
& 82.61 & \textbf{55.02} & \textbf{51.01} & \textbf{0.039} & \textbf{0.706} 
\\
 \midrule
 AdaAD & 84.51& 50.67 & 46.56 & 0.069 & 0.516 
& 84.24 & 55.47& 51.36& 0.034 & 0.777 
\\
 AdaIAD & 85.04& 51.33& 47.62& 0.068& 0.519& 84.14& 55.58& 51.48& 0.034& 0.778\\
 \textbf{+IGDM} & 84.75& \textbf{52.35} & \textbf{48.34} & \textbf{0.065} & \textbf{0.526} & 84.27& \textbf{56.15} & \textbf{52.15} & \textbf{0.033} & \textbf{0.790} \\

\bottomrule
 \end{tabular}
 \label{tab:MNV2_cifar10}
\end{table*}

In \Cref{tab:MNV2_cifar100} and \Cref{tab:MNV2_cifar10}, we present additional experiments on the CIFAR-10 and CIFAR-100 datasets using MobileNetV2 \citep{sandler2018mobilenetv2} as the student model, with two distinct teacher models: LTD and BDM-AT. Across both datasets, applying IGDM consistently improved robustness under PGD and AutoAttack, compared to the baseline adversarial training methods. 
Notably, IGDM enhanced the model’s performance on both CIFAR-10 and CIFAR-100, particularly in reducing the Gradient Distance (GD) and improving Gradient Cosine similarity (GC). These results underscore the effectiveness of IGDM in improving adversarial robustness and gradient alignment across different datasets and architectures.

\subsection{Adversarial Distillation with IKL-AT Teacher Model.}
\label{sec:IKL-AT}
\begin{table*}[h]
 \centering
 \caption{Adversarial distillation results on ResNet-18 and MobileNetV2 with a IKL-AT teacher model on CIFAR-100. Bold indicates cases where IGDM improved accuracy by more than 0.5 percentage points or better gradient matching.}
 \setlength{\tabcolsep}{6pt}
 \begin{tabular}{lcccccccccc}
 \toprule
    \multicolumn{1}{c}{\multirow{2.5}{*}{Method}}  & \multicolumn{5}{c}{CIFAR-100 with ResNet-18 student}& \multicolumn{5}{c}{CIFAR-100 with MobileNetV2 student} \\
   \cmidrule(r){2-6}
   \cmidrule(r){7-11}
   & Clean & PGD & AA & GD$\Downarrow$ & GC$\Uparrow$ & Clean & PGD & AA & GD$\Downarrow$ & GC$\Uparrow$ \\
\midrule

 PGD-AT  & 55.80 & 19.88 & 18.86 & 0.460 & 0.306 & 59.23 & 24.04 & 21.58 & 0.211 
& 0.392
\\
 \textbf{+IGDM} & \textbf{62.91} & \textbf{35.25} & \textbf{28.89} & \textbf{0.122}& \textbf{0.534} & \textbf{59.75} & \textbf{33.59} & \textbf{27.02} & \textbf{0.120} 
& 
\textbf{0.517}
\\
 \midrule
 TRADES & 53.56 & 25.85 & 22.02 & 0.199& 0.354 & 51.05 & 24.83 & 20.62 & 0.148 
& 0.351
\\
 \textbf{+IGDM}  & \textbf{62.41} & \textbf{31.79} & \textbf{24.02} & \textbf{0.116}& \textbf{0.467} & \textbf{59.34} & \textbf{29.36} & \textbf{22.40} & \textbf{0.120} 
& 
\textbf{0.444}
\\
 \midrule
 ARD & 61.38 & 27.59 & 23.18 & 0.178& 0.400 & 61.52 & 28.00 & 23.40 & 0.164 
& 0.420 
\\
 \textbf{+IGDM} & 61.55 & \textbf{35.24} & \textbf{28.87} & \textbf{0.117}& \textbf{0.546} & 61.95 & \textbf{32.26}& \textbf{26.48} & \textbf{0.132} 
& 
\textbf{0.498} 
\\
 \midrule

 IAD & 61.09 & 29.25 & 23.61 & 0.173& 0.403 & 58.39 & 28.40 & 23.06 & 0.147 
& 0.418
\\
 \textbf{+IGDM} & \textbf{64.20} & \textbf{33.71} & \textbf{27.03} & \textbf{0.132}& \textbf{0.506} & \textbf{59.58} & \textbf{30.96} & \textbf{25.42} & \textbf{0.128} 
& 
\textbf{0.478}
\\
 \midrule
 AKD & 61.90& 29.23& 23.96& 0.171& 0.412& 61.69& 27.87& 23.60& 0.159& 0.431\\
 \textbf{+IGDM} & 61.94& \textbf{33.16}& \textbf{27.82}&\textbf{ 0.131}& \textbf{0.511}& 
 61.56 & \textbf{33.62} & \textbf{27.32} & \textbf{0.117} & \textbf{0.519} 
\\
 \midrule
 RSLAD & 61.18 & 30.54 & 25.27 & 0.147&0.444 & 61.95 & 30.16 & 25.09 & 0.157 
& 0.441\\
 \textbf{+IGDM} & \textbf{63.55} & \textbf{35.26} & \textbf{29.10} & \textbf{0.119} & \textbf{0.541} & 
\textbf{62.48} & \textbf{35.25} & \textbf{28.82} & 
\textbf{0.118} 
& 
\textbf{0.547}\\
 \midrule
 AdaAD &65.36& 32.29 & 26.89 & 0.133& 0.525 & 62.35 & 28.51 & 23.01 & 0.140 
& 0.470\\
 AdaIAD & 65.29& 32.47& 26.80& 0.133& 0.527& 
62.17& 28.48& 23.39& 0.139& 
0.473\\
 \textbf{+IGDM} & \textbf{66.00} & \textbf{34.47} & \textbf{29.22} & \textbf{0.119}& \textbf{0.582} & 62.02 & \textbf{32.85} & \textbf{26.11} & \textbf{0.119} & \textbf{0.533}\\
\bottomrule
 \end{tabular}
 \label{tab:ikl_at_cifar100}
  \vspace{-0.5cm}
\end{table*}

\begin{table*}[h]
 \centering
 \caption{Adversarial distillation results on ResNet-18 and MobileNetV2 with a IKL-AT teacher model on CIFAR-10. Bold indicates cases where IGDM improved accuracy by more than 0.5 percentage points or better gradient matching.}
 \setlength{\tabcolsep}{6pt}
 \begin{tabular}{lcccccccccc}
 \toprule
   \multicolumn{1}{c}{\multirow{2.5}{*}{Method}}& \multicolumn{5}{c}{CIFAR-10 with ResNet-18 student}& \multicolumn{5}{c}{CIFAR-10 with MobileNetV2 student} \\
   \cmidrule(r){2-6}
   \cmidrule(r){7-11}
   & Clean & PGD & AA & GD$\Downarrow$ & GC$\Uparrow$ & Clean & PGD & AA & GD$\Downarrow$ & GC$\Uparrow$ \\
\midrule PGD-AT  & 84.52 & 42.80 & 41.12 & 0.409 & 0.286 & 83.52 & 44.47 & 41.19 & 0.175 
& 0.401 
\\
 \textbf{+IGDM} & 83.66 & \textbf{54.38} & \textbf{49.85} & \textbf{0.072} & \textbf{0.541} & 82.11 & \textbf{51.13} & \textbf{47.00} & \textbf{0.063} 
& 
\textbf{0.507}
\\
\midrule TRADES  & 82.46 & 49.13 & 47.09 & 0.131 & 0.425 & 81.57 & 50.49 & 46.88 & 0.073 
& 0.489
\\
 \textbf{+IGDM} &\textbf{83.14} & \textbf{54.91} & \textbf{48.76} &\textbf{0.098} & \textbf{0.455} &81.67 & \textbf{52.32} & \textbf{47.40} & \textbf{0.057} 
& 
\textbf{0.492}
\\
\midrule
 ARD & 85.41 & 49.36 & 45.32 & 0.087 & 0.389 & 84.68 & 48.03 & 44.11 & 0.087 
& 0.376 
\\
 \textbf{+IGDM} & 85.78 & \textbf{54.37} & \textbf{46.52} & \textbf{0.072} & \textbf{0.467} & 84.72 & \textbf{50.89} & \textbf{46.13} & \textbf{0.070} 
& 
\textbf{0.473} 
\\
 \midrule
 IAD & 85.22 & 49.70 & 45.96 & 0.083 & 0.393 & 84.25 & 48.58 & 44.15 & 0.082 
& 0.390 
\\
 \textbf{+IGDM} & 83.78 & \textbf{54.77} & \textbf{50.24} & \textbf{0.061} & \textbf{0.521} & 83.89 & \textbf{51.46} & \textbf{46.48} & \textbf{0.069} 
&
\textbf{0.458} 
\\
 \midrule
 AKD & 83.62& 51.81& 46.01& 0.080& 0.395& 84.52 & 47.56 & 43.27 & 0.154 & 0.421 \\
 \textbf{+IGDM} & 83.70& \textbf{53.69}& \textbf{49.45}& \textbf{0.067}& \textbf{0.539}& 84.33 & \textbf{50.36} & \textbf{45.37} & \textbf{0.089} & \textbf{0.438} 
\\
 \midrule
 RSLAD & 85.55 & 52.13 & 48.83 & 0.069 & 0.531 & 84.62& 51.43 & 47.69 & 0.072 
& 0.503\\
 \textbf{+IGDM} & 85.67& \textbf{53.30} & \textbf{49.58} & \textbf{0.065} & \textbf{0.541} & 84.30& \textbf{53.14} & \textbf{48.91} & 
\textbf{0.066} 
& 
\textbf{0.523}\\
 \midrule
 AdaAD & 86.59& 54.25 & 50.86 & 0.062 & 0.570 & 85.31 & 52.10 & 48.03 & 0.066 
& 0.524\\
 AdaIAD & 86.57& 54.59& 50.61& 0.062& 0.572& 85.47& 52.05& 47.96& 0.066& 
0.523\\
 \textbf{+IGDM} & 86.10 & \textbf{55.41} & \textbf{51.29} & \textbf{0.060} & \textbf{0.587} & 85.11& \textbf{53.07} & \textbf{49.08} & \textbf{0.061} & \textbf{0.541}\\

\bottomrule
 \end{tabular}
 \label{tab:ikl_at_cifar10}
\end{table*}
In \Cref{tab:ikl_at_cifar100} and \Cref{tab:ikl_at_cifar10}, we conduct additional experiments on the CIFAR-100 and CIFAR-10 datasets with the IKL-AT \citep{cui2023decoupled} teacher model.
Similar to the outcomes observed with the LTD \citep{LTD} and BDM-AT \citep{wang2023better} teachers on those datasets, IGDM demonstrates substantial enhancements in robustness with the IKL-AT teacher, regardless of the original methods, datasets, or student models, while maintaining consistent clean accuracy.
Furthermore, consistent results are observed across all experiments, with a decrease in GD and an increase in GC upon applying IGDM.

\subsection{Adversarial Distillation on Tiny-ImageNet}
\label{sec:diff_dataset}

\begin{table}[h]
 \centering
\caption{Adversarial distillation result on PreActResNet-18 with a WideResNet-34-10 teacher model on Tiny-ImageNet. Bold indicates cases where IGDM improved accuracy by more than 0.5 percentage points or better gradient matching.}
 \setlength{\tabcolsep}{13.5pt}
 \begin{tabular}{lccccc}
 \toprule
 \multicolumn{1}{c}{Method} & Clean & PGD & AA & GD$\Downarrow$ & GC$\Uparrow$  \\
\midrule
PGD-AT \citep{PGD}& 50.13 & 17.10 & 14.58 & 0.126 & 0.490 \\
\textbf{+IGDM} & \textbf{51.46} & \textbf{25.04} & \textbf{20.10} & \textbf{0.071} & \textbf{0.672} \\
\midrule
TRADES \citep{TRADES}& 46.51 & 19.51 & 14.89 & 0.094 & 0.447 \\
\textbf{+IGDM} & \textbf{51.31} & \textbf{24.10} & \textbf{18.92} & \textbf{0.073} & \textbf{0.649} \\
\midrule
ARD \citep{ard}& 50.81 & 23.37 & 19.48 & 0.081 & 0.613 \\
\textbf{+IGDM} & 50.86 & \textbf{25.39} & \textbf{20.57} & \textbf{0.069} & \textbf{0.683} \\
\midrule
IAD \citep{iad}& 49.89 & 23.89 & 19.10 & 0.079 & 0.614 \\
\textbf{+IGDM} & 50.02 & \textbf{25.57} & \textbf{20.69} & \textbf{0.066} & \textbf{0.704} \\
\midrule
AKD \citep{akd}& 51.70& 23.86& 19.69& 0.082& 0.612\\
\textbf{+IGDM} & 51.38& \textbf{25.25}& \textbf{20.99}& \textbf{0.063}& \textbf{0.720}\\
 \midrule
RSLAD \citep{rslad}& 47.12 & 22.14 & 17.65 & 0.081 & 0.567 \\
\textbf{+IGDM} & 47.54 & \textbf{23.95} & \textbf{18.35} & \textbf{0.074} & \textbf{0.620} \\
\midrule
AdaAD \citep{adaad}& 51.54 & 24.65 & 20.60 & 0.064 & 0.738 \\
\textbf{+IGDM} & 51.21 & \textbf{25.73} & \textbf{21.17} & \textbf{0.060} & \textbf{0.764} \\
\bottomrule
 \end{tabular}

 \label{tab:Tiny}
\end{table}
The results in \Cref{tab:Tiny} confirm the effectiveness of IGDM in enhancing adversarial robustness, especially on the challenging Tiny-ImageNet dataset \cite{le2015tiny}. 
Consistent with findings on other datasets, IGDM improves robustness metrics such as PGD and AA accuracy, while maintaining comparable clean accuracy. This improvement is particularly significant given the complexity of Tiny-ImageNet, where achieving high robustness is often challenging. 
Moreover, IGDM facilitates better alignment between teacher and student models, evidenced by a reduction in mean Gradient Distance (GD) and an increase in mean Gradient Cosine similarity (GC), further underscoring the impact of gradient matching of our method.

\subsection{Drawbacks of Direct Gradient Matching in Contrast to IGDM}
\label{sec:direct_matching}
\begin{table}[h]
 \centering
 \caption{Adversarial distillation result of ARD variant methods on ResNet-18 with BDM-AT teacher on CIFAR-100. Direct$_{\text{ARD}}$ represents the direct distillation of gradients, and the $\alpha$ is the hyperparameter of the gradient distillation loss.}
 \setlength{\tabcolsep}{14pt}
 \begin{tabular}{lccccc}
 \toprule
 \multicolumn{1}{c}{Method}  & Clean & PGD & AA & GD$\Downarrow$ & GC$\Uparrow$ \\
\midrule

% IGDM$_{\text{ARD}} (\alpha = 1)$  & 60.56 & 31.55 & 26.02 & 0.130& 0.467\\ 
% IGDM$_{\text{ARD}} (\alpha = 5)$  & 59.72 & 33.02 & 27.32 & 0.122& 0.494\\ 
% IGDM$_{\text{ARD}} (\alpha = 10)$  & 59.72 & 33.79 & 28.07 & 0.116 & 0.512 \\
% %IGDM$_{\text{ARD}} (\alpha = 20)$  & 0 & 0 & 0 & 0 & 0 \\
% IGDM$_{\text{ARD}} (\alpha = 30)$  & 60.35 & 35.02 & 28.93 & 0.106 & 0.545 \\
% %IGDM$_{\text{ARD}} (\alpha = 40)$  & 60.36 & 35.07 & 28.91 & 0 & 0 \\
% %IGDM$_{\text{ARD}} (\alpha = 50)$  & 60.56 & 34.89 & 29.51 & 0.102& 0.563\\
% IGDM$_{\text{ARD}} (\alpha = 60)$  & 60.28 & 35.75 & 29.42 & 0.100& 0.570\\
% IGDM$_{\text{ARD}} (\alpha = 100)$  & 59.98 & 35.47 & 29.5 & 0.098& 0.582\\
% \midrule
Direct$_{\text{ARD}} (\alpha = 1)$ & 60.92 & 30.56 & 25.23 & 0.136 & 0.447 \\
Direct$_{\text{ARD}} (\alpha = 3)$ & 60.98 & 30.29 & 24.88 & 0.137 & 0.444 \\
Direct$_{\text{ARD}} (\alpha = 6)$ & 61.13 & 30.34 & 25.27 & 0.135 & 0.448 \\
%Direct$_{\text{ARD}} (\alpha = 5)$ & 60.94 & 30.53 & 25.25 & 0.134 & 0.453 \\
Direct$_{\text{ARD}} (\alpha = 10)$ & 61.05 & 30.80 & 25.33 & 0.132 & 0.453 \\
Direct$_{\text{ARD}} (\alpha = 30)$ & 61.43 & 30.45 & 25.29 & 0.127 & 0.455 \\
Direct$_{\text{ARD}} (\alpha = 60)$ & 61.14 & 30.17 & 24.73 & 0.124 & 0.450 \\
Direct$_{\text{ARD}} (\alpha = 100)$ & 60.36 & 30.48 & 24.62 & 0.121 & 0.445 \\
%Direct$_{\text{ARD}} (\alpha = 200)$ & 60.62 & 30.16 & 24.22 & 0.118 & 0.442 \\
Direct$_{\text{ARD}} (\alpha = 300)$ & 59.63 & 29.72 & 23.99 & 0.116 & 0.440 \\
Direct$_{\text{ARD}} (\alpha = 600)$ & 58.79 & 29.76 & 24.06 & 0.111 & 0.441 \\
Direct$_{\text{ARD}} (\alpha = 1000)$ & 57.08 & 29.89 & 24.14 & 0.108 & 0.445 \\
Direct$_{\text{ARD}} (\alpha = 3000)$ & 51.30 & 29.13 & 22.85 & 0.104 & 0.441 \\
Direct$_{\text{ARD}} (\alpha = 6000)$ & 46.95 & 27.33 & 20.70 & 0.105 & 0.424 \\
%Direct$_{\text{ARD}} (\alpha = 5000)$ & 48.08 & 28.04 & 21.58 & 0.104 & 0.431 \\
Direct$_{\text{ARD}} (\alpha = 10000)$ & 43.22 & 25.67 & 19.18 & 0.105 & 0.404 \\
\midrule
 ARD \citep{ard} & 61.51 & 30.23 & 24.77 & 0.142& 0.439\\
 \textbf{+IGDM} & \textbf{61.62} & \textbf{35.75} & \textbf{28.79} & \textbf{0.101}& \textbf{0.571}\\

\bottomrule
 \end{tabular}
 \label{tab:supp_direct_alpha}
\end{table}
IGDM achieves gradient matching indirectly by leveraging the differences in logits. One might assume that distilling directly calculated gradients could achieve gradient matching more intuitively, as expressed by the following equation:
\begin{equation}
    L_{Direct} = T(\alpha) \cdot D\left (\frac{\partial f_S(\textbf{x})}{\partial \textbf{x}} \ , \ \frac{\partial f_T(\textbf{x})}{\partial \textbf{x}} \right ) .
\end{equation}
In \Cref{tab:supp_direct_alpha}, we conduct experiment on IGDM$_{\text{ARD}}$ and direct gradient matching (Direct$_{\text{ARD}}$).
To ensure a fair comparison between the two methods, we kept all other factors the same, varying only the gradient calculation method (logits differences vs. direct calculation).
The results reveal that direct gradient distillation fails to meaningfully enhance robustness compared to the original ARD method.
Although it shows a slight enhancement in gradient matching, its performance significantly lags behind IGDM.

\begin{figure}[h]
\centering
     \includegraphics[width=\textwidth] {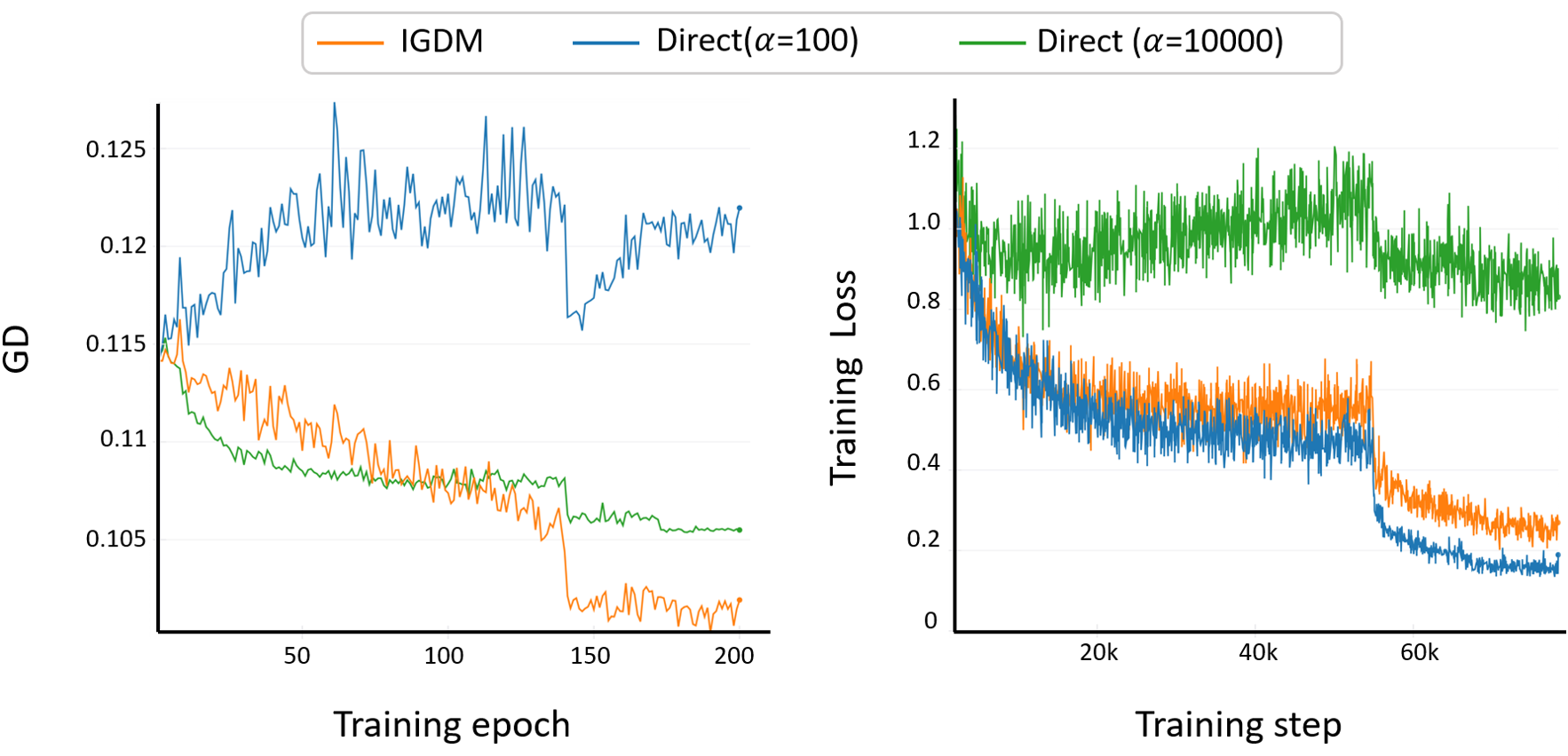}%
     \caption{Comparision between IGDM and direct gradient matching. GD and training loss are measured at the training epoch and step, respectively.}
     \label{fig:direct_igdm}
\end{figure}

To provide a comprehensive interpretation, we analyze GD and training loss over training time, epoch, and step in \Cref{fig:direct_igdm}.
The limitations of direct gradient matching are evident due to optimization challenges; a small $\alpha$ fails to match the gradient, while a large $\alpha$ results in poor convergence of the training loss. 
More interestingly, even with a large $\alpha$, direct matching results in GD values higher than those achieved by IGDM.
We interpret these optimization issues as arising from the low-level features of the input gradient, similar to the input itself.
In other words, directly matching low-level feature gradients is comparable to matching pixel-wise image differences. 
This leads to the model's inability to effectively capture the training objective, failing to match the gradient and, consequently, diminished robustness.
One potential solution to address these optimization challenges is to employ an additional discriminator model, as proposed in \citep{chan2020thinks}.
However, this approach requires training an additional discriminator model, whereas IGDM does not necessitate any other model or training procedure.
In summary, IGDM significantly enhances robustness by indirectly matching the gradient using high-level feature logits, thereby achieving superior robustness.
%Moreover, higher hyperparameter values contribute to better gradient matching, resulting in improved robustness.

\subsection{Important Role of Logit Difference in IGDM}
Adversarial training aims to make the network output almost constant within a ball around the input point.
Therefore, one might assume that the distilling logit difference of the adversarially trained teacher model in IGDM loss is close to distilling 0.
In other words, IGDM can be simplified as expressed by the following equation:
\begin{equation}
\begin{split}
    L_{\text{IGDM}_\text{TRADES-like}} &=  T(\alpha) \cdot D (f_S(\textbf{x} +  \boldsymbol{\delta} ) - f_S(\textbf{x} -  \boldsymbol{\delta} ) \ , \ \mathbf{0}) \\
     &= T(\alpha) \cdot D (f_S(\textbf{x} +  \boldsymbol{\delta} ) \ ,\   f_S(\textbf{x} -  \boldsymbol{\delta} )).
\end{split}
\end{equation}
We refer to this method as IGDM$_\text{TRADES-like}$ because
it resembles a TRADES \citep{TRADES} regularization loss (TRADES$_\text{reg}$), and empirically, its performance closely aligns with the addition of the TRADES regularization loss:
\begin{equation}
    L_{\text{TRADES}_\text{reg}} =  T(\alpha) \cdot D (f_S(\textbf{x} + \boldsymbol{\delta} ) \,\ f_S(\textbf{x})).
\end{equation}
\label{sec:logit_diff}
\begin{table}[h]
 \centering
 \caption{Adversarial distillation result of ARD variant methods on ResNet-18 with BDM-AT teacher on CIFAR-100. TRADES$_\text{reg}$ represents the regularization loss of TRADES. and the $\alpha$ is the hyperparameter of the gradient distillation loss.}
 \setlength{\tabcolsep}{12.3pt}
 \begin{tabular}{lccccc}
 \toprule

 \multicolumn{1}{c}{Method}  & Clean & PGD & AA & GD$\Downarrow$ & GC$\Uparrow$ \\
\midrule

ARD $+$ TRADES$_{\text{reg}} (\alpha = 1)$  & 57.86 & 31.44 & 25.95 & 0.125 & 0.459\\
ARD $+$ TRADES$_{\text{reg}} (\alpha = 5)$  & 58.07 & 32.13 & 26.06 & 0.117 & 0.466\\
ARD $+$ TRADES$_{\text{reg}} (\alpha = 10)$  & 53.58 & 32.01 & 26.27 & 0.116 & 0.463\\
ARD $+$ TRADES$_{\text{reg}} (\alpha = 15)$  & 52.05 & 31.53 & 26.23 & 0.115 & 0.464\\
ARD $+$ TRADES$_{\text{reg}} (\alpha = 20)$  & 50.87 & 31.92 & 26.30 & 0.115 & 0.462\\
ARD $+$ TRADES$_{\text{reg}} (\alpha = 25)$  & 50.28 & 31.92 & 26.15 & 0.115 & 0.458\\
ARD $+$ TRADES$_{\text{reg}} (\alpha = 30)$  & 49.20 & 31.93 & 25.87 & 0.115 & 0.456\\
\midrule
ARD $+$ IGDM$_\text{TRADES-like} (\alpha = 1)$  & 57.73 & 31.72 & 25.64 & 0.119 & 0.461\\
ARD $+$ IGDM$_\text{TRADES-like} (\alpha = 5)$  & 54.18 & 31.71 & 25.79 & 0.117& 0.466\\
ARD $+$ IGDM$_\text{TRADES-like} (\alpha = 10)$  & 52.21 & 32.08 & 25.98 & 0.118& 0.464\\
ARD $+$ IGDM$_\text{TRADES-like} (\alpha = 15)$  & 50.50 & 31.85 & 25.91 & 0.118& 0.457\\
ARD $+$ IGDM$_\text{TRADES-like} (\alpha = 20)$  & 49.67 & 31.76 & 25.78 & 0.118& 0.452\\
ARD $+$ IGDM$_\text{TRADES-like} (\alpha = 25)$  & 48.21 & 31.33 & 25.38 & 0.119& 0.445\\
ARD $+$ IGDM$_\text{TRADES-like} (\alpha = 30)$  & 47.61 & 31.04 & 25.05 & 0.119& 0.439\\
\midrule
 ARD \citep{ard} & 61.51 & 30.23 & 24.77 & 0.142& 0.439\\
 \textbf{+IGDM} & \textbf{61.62} & \textbf{35.75} & \textbf{28.79} & \textbf{0.101}& \textbf{0.571}\\

\bottomrule
 \end{tabular}

 \label{tab:supp_logit_diff}
  % \vspace*{-0.3cm}
\end{table}

In \Cref{tab:supp_logit_diff}, we conduct experiments on these two methods in conjunction with the ARD \citep{ard} method.
For TRADES$_\text{reg}$, as the regularization term strengthens, the robustness slightly increases; however, it lags far behind IGDM in terms of robustness. 
IGDM$_\text{TRADES-like}$ follows a similar trend to the TRADES$_\text{reg}$ method with the increase in hyperparameter.
These results indicate that the logit difference of the teacher model provides valuable information for teaching the student model to achieve robustness through gradient matching, resulting in superior robustness. This finding corroborates our main contribution in the main paper.

\subsection{Analysis on Hyperparameter Sensitivity of IGDM}
\label{sec:hyperparameter}
The proposed IGDM loss is defined as follows:
\begin{equation*}
L_{IGDM} = T(\alpha) \cdot D\left (f_S(\textbf{x} + \boldsymbol{\delta} ) - f_S(\textbf{x} - \boldsymbol{\delta} )\ , \ f_T(\textbf{x} + \boldsymbol{\delta}) - f_T(\textbf{x} - \boldsymbol{\delta} ) \right ).
\end{equation*}
In \Cref{tab:supp_hyper_pgd} and \Cref{tab:supp_hyper_ard}, we analyze the sensitivity of hyperparameters $\alpha$ using IGDM$_\text{PGD-AT}$ and IGDM$_\text{ARD}$.
In these tables, a value of 0 for $\alpha$ represents the original methods, PGD-AT and ARD, respectively.
As $\alpha$ increases, gradient matching becomes more prominent, leading to enhanced robustness for both methods.
However, excessively large values of $\alpha$ do not provide additional enhancements and may even reduce robustness.
%Conversely, values in the range from 1.0 to 1.5 for $\beta$ demonstrate particularly strong gradient matching, resulting in strong robustness.
Based on the insights from the hyperparameter sensitivity analysis, we employed grid search to adapt the IGDM module for all experiments.

\begin{table}[h]
 \centering
 \caption{Analysis of hyperparameter of IGDM using PGD-AT variant methods on ResNet-18 with BDM-AT teacher on CIFAR-100. $\alpha$ is the hyperparameter of IGDM loss. The gray row indicates the reported values.}
 \setlength{\tabcolsep}{14pt}
 \begin{tabular}{lccccc}
 \toprule

 Method& Clean & PGD & AA & GD$\Downarrow$ & GC$\Uparrow$ \\
\midrule
 PGD-AT \citep{PGD}& 55.80 & 19.88 & 18.86 & 0.452 & 0.389 \\
%IGDM$_\text{PGD-AT}$ & 1 & 1 & 56.15 & 23.70 & 21.38 & 0.256 & 0.362 \\
%IGDM$_\text{PGD-AT}$ & 5 & 1 & 56.45 & 26.19 & 23.00 & 0.198 & 0.389\\
%10 & 1 & 56.72 & 26.89 & 23.28 & 0.178 & 0.409\\
 IGDM$_\text{PGD-AT}$($\alpha=20)$  & 56.95 & 28.10 & 24.02 & 0.157 & 0.427\\
%IGDM$_\text{PGD-AT}$ & 30 & 1 & 57.21 & 29.38 & 25.19 & 0& 0\\
 IGDM$_\text{PGD-AT}$($\alpha=40)$ & 58.23 & 30.36 & 25.71 & 0.139 & 0.457\\
%IGDM$_\text{PGD-AT}$ & 50 & 1 & 58.58 & 30.96 & 25.92 & 0& 0\\
 IGDM$_\text{PGD-AT}$($\alpha=60)$& 58.74 & 31.69 & 26.57 & 0.131 & 0.483\\
%IGDM$_\text{PGD-AT}$ & 70 & 1 & 58.58 & 32.08 & 26.79 & 0& 0\\
 IGDM$_\text{PGD-AT}$($\alpha=80)$& 59.01 & 32.02 & 27.08 & 0.123 & 0.504\\
%IGDM$_\text{PGD-AT}$ & 90 & 1 & 59.05 & 32.61 & 27.16 & 0& 0\\
 IGDM$_\text{PGD-AT}$($\alpha=100)$& 59.20 & 33.09 & 27.48 & 0.118 & 0.519\\
 IGDM$_\text{PGD-AT}$($\alpha=120)$& 59.42 & 33.60 & 27.95 & 0.114 & 0.531\\
 IGDM$_\text{PGD-AT}$($\alpha=140)$& 59.32 & 34.18 & 28.26 & 0.111 & 0.542\\
 IGDM$_\text{PGD-AT}$($\alpha=160)$& 60.06 & 34.38 & 28.47 & 0.108 & 0.550\\
 IGDM$_\text{PGD-AT}$($\alpha=180)$& 60.53 & 34.47 & 28.51 & 0.105 & 0.556\\
 IGDM$_\text{PGD-AT}$($\alpha=200)$& 60.83 & 34.90 & 28.84 & 0.104 & 0.562\\
 IGDM$_\text{PGD-AT}$($\alpha=220)$& 60.68 & 35.09 & \textbf{29.16} & 0.102 & 0.569\\
 IGDM$_\text{PGD-AT}$($\alpha=240)$& 59.65 & 35.34 & 29.10 & 0.101 & 0.572\\
 IGDM$_\text{PGD-AT}$($\alpha=260)$& 59.71 & 35.37 & 29.09 & 0.100 & 0.573\\
 IGDM$_\text{PGD-AT}$($\alpha=280)$& 59.83 & 35.23 & 28.84 & 0.099 & 0.576\\
 IGDM$_\text{PGD-AT}$($\alpha=300)$& 59.95 & 35.39 & 29.13 & 0.098 & 0.579\\
%IGDM$_\text{PGD-AT}$ & 100 & 0.1 & 56.58 & 24.07 & 21.78 & 0.241 & 0.372\\
%IGDM$_\text{PGD-AT}$ & 100 & 0.2 & 56.94 & 25.48 & 22.44 & 0& 0\\
%100 & 0.3 & 56.88 & 26.88 & 23.11 & 0.177 & 0.403\\
%1000 & 0.4 & 57.21 & 27.58 & 23.39 & 0.162 & 0.424\\
% 100 & 0.5 & 57.61 & 28.91 & 24.16 & 0.153 & 0.439\\
% 100 & 0.6 & 57.85 & 29.93 & 25.06 & 0.144& 0.455\\
% 100 & 0.7 & 58.27 & 30.77 & 25.80 & 0.136 & 0.475\\
% 100 & 0.8 & 58.65 & 31.45 & 26.27 & 0.129& 0.491\\
% 100 & 0.9 & 58.96 & 32.55 & 26.96 & 0.123 & 0.505\\
% 100 & 1.0 & 59.20 & 33.09 & 27.48 & 0.118 & 0.519\\
% 100 & 1.1 & 60.19 & 33.72 & 27.87 & 0.114 & 0.529\\
% 100 & 1.2 & 59.71 & 33.98 & 28.31 & 0.111& 0.540\\
% 100 & 1.3 & 61.00 & 34.60 & 28.60 & 0.109 & 0.548\\
% 100 & 1.4 & 60.95 & 34.70 & 28.78 & 0.107& 0.555\\
% 100 & 1.5 & 61.86 & 34.14 & 27.52 & 0.106 & 0.529\\
% 100 & 1.6 & 61.01 & 32.52 & 24.87 & 0.105& 0.469\\
% 100 & 1.7 & 59.67 & 31.81 & 24.51 & 0.105 & 0.454\\
% 100 & 1.8 & 59.16 & 31.19 & 23.40 & 0.106 & 0.430\\
% 100 & 1.9 & 60.82 & 32.00 & 24.39 & 0.104 & 0.468\\
% 100 & 2.0 & 60.35 & 31.53 & 24.39 & 0.104 & 0.471\\
\bottomrule
 \end{tabular}

 \label{tab:supp_hyper_pgd}
  % \vspace*{-0.3cm}
\end{table}

\begin{table}[h]
 \centering
 \caption{Analysis of hyperparameter of IGDM using ARD variant methods on ResNet-18 with BDM-AT teacher on CIFAR-100. $\alpha$ is the hyperparameter of IGDM loss. The gray row indicates the reported values.}
 \setlength{\tabcolsep}{14pt}
 \begin{tabular}{lccccc}
 \toprule

 Method& Clean & PGD & AA & GD$\Downarrow$ & GC$\Uparrow$ \\
\midrule
 ARD \citep{ard}& 61.51 & 30.23 & 24.77 & 0.1422 & 0.439 
\\
%IGDM$_\text{PGD-AT}$ & 1 & 1 & 56.15 & 23.70 & 21.38 & 0.256 & 0.362 \\
%IGDM$_\text{PGD-AT}$ & 5 & 1 & 56.45 & 26.19 & 23.00 & 0.198 & 0.389\\
%10 & 1 & 56.72 & 26.89 & 23.28 & 0.178 & 0.409\\
 IGDM$_\text{ARD}$($\alpha=20)$  & 59.74 & 33.97 & 27.85 & 0.112 & 0.523
\\
%IGDM$_\text{PGD-AT}$ & 30 & 1 & 57.21 & 29.38 & 25.19 & 0& 0\\
 IGDM$_\text{ARD}$($\alpha=40)$ & 60.13& 34.21& 28.56& 0.106 & 0.546
\\
%IGDM$_\text{PGD-AT}$ & 50 & 1 & 58.58 & 30.96 & 25.92 & 0& 0\\
 IGDM$_\text{ARD}$($\alpha=60)$& 61.08& 34.54& 28.42& 0.104& 0.557\\
%IGDM$_\text{PGD-AT}$ & 70 & 1 & 58.58 & 32.08 & 26.79 & 0& 0\\
 IGDM$_\text{ARD}$($\alpha=80)$& 60.78& 35.21& 28.73& 0.102& 0.568\\
%IGDM$_\text{PGD-AT}$ & 90 & 1 & 59.05 & 32.61 & 27.16 & 0& 0\\
 IGDM$_\text{ARD}$($\alpha=100)$& 61.18& 35.75& \textbf{28.79}& 0.101& 0.571\\
 IGDM$_\text{ARD}$($\alpha=120)$& 61.15& 35.17& 28.67& 0.099& 0.578\\
 IGDM$_\text{ARD}$($\alpha=140)$& 61.24& 35.32& 28.75& 0.098& 0.583\\
 IGDM$_\text{ARD}$($\alpha=160)$& 60.76& 35.81& 28.71& 0.096 & 0.584\\
 IGDM$_\text{ARD}$($\alpha=180)$& 60.19& 35.67& 28.78& 0.095 & 0.585\\
 IGDM$_\text{ARD}$($\alpha=200)$& 59.95& 35.19 & 28.67& 0.095 & 0.585
\\
 IGDM$_\text{ARD}$($\alpha=220)$& 59.42& 35.24& 28.76& 0.096 & 0.588\\
 IGDM$_\text{ARD}$($\alpha=240)$& 59.76& 34.78& 28.74& 0.095 & 0.587\\
 IGDM$_\text{ARD}$($\alpha=260)$& 59.04 & 35.01 & 28.36 & 0.095 & 0.586
\\
 IGDM$_\text{ARD}$($\alpha=280)$& 59.56 & 35.21 & 28.59 & 0.095 & 0.589
\\
 IGDM$_\text{ARD}$($\alpha=300)$& 58.91 & 34.78 & 28.58 & 0.095 & 0.582\\
\bottomrule
 \end{tabular}

 \label{tab:supp_hyper_ard}
\end{table}

\section{Training details}
\label{sec:apendix_training_detail}
\subsection{Settings}
We utilized the CIFAR-10, CIFAR-100 \citep{krizhevsky2009learning}, SVHN \cite{svhn}, and Tiny-ImageNet \citep{le2015tiny} datasets with random crop and random horizontal flips, excluding other augmentations.
We trained all AT, AD, and IGDM incorporated methods using an SGD momentum optimizer with the same initial learning rate of 0.1, momentum of 0.9, and weight decay of 5e-4. 

For CIFAR-10 and CIFAR-100, we adhered to the training settings of other adversarial distillation methods, training for 200 epochs, except for RSLAD and IGDM$_{\text{RSLAD}}$, which were trained for 300 epochs. 
RSLAD suggested that increasing the number of training epochs could enhance model robustness; thus, we followed their recommendation to train for 300 epochs.
The learning rate scheduler reduced the learning rate by a factor of 10 at the 100th and 150th epochs. However, for RSLAD and IGDM$_{\text{RSLAD}}$, the learning rate decreased by 10 at the 215th, 260th, and 285th epochs, as suggested in the original paper. 
For SVHN, we trained for 50 epochs with the learning rate decayed by a factor of 10 at the 40th and 45th epochs for all methods. For Tiny-ImageNet, we trained for 100 epochs with the learning rate decayed by a factor of 10 at the 50th and 80th epochs for all methods.

The adversarial perturbation settings were as follows: the number of iterations for inner maximization was set to 10, with a step size of $2/255$, and a total perturbation bound of $L_\infty = 8/255$. 
Specifically, we employed the recommended inner loss functions as outlined in the original paper: PGD attack on student model for PGD-AT, ARD, and IAD; TRADES attack on student model for TRADES; RSLAD inner loss for RSLAD; and AdaAD inner loss for AdaAD.
Moreover, the IGDM-incorporated method followed the inner maximization method of the original AT or AD method.
For formulaic representation as follows,
\begin{align}
    \text{PGD attack: } & \boldsymbol{\delta} = \argmax_{ \| \boldsymbol{\delta'} \|_p \leq \epsilon} \operatorname{CE} (f_S  (\mathbf{x}+\boldsymbol{\delta'}), y), \\
    \text{TRADES attack: } & \boldsymbol{\delta} = \argmax_{ \| \boldsymbol{\delta'} \|_p \leq \epsilon} \operatorname{KL} (f_S  (\mathbf{x}+\boldsymbol{\delta'}), f_S  (\mathbf{x})), \\
    \text{RSLAD attack: } & \boldsymbol{\delta} = \argmax_{ \| \boldsymbol{\delta'} \|_p \leq \epsilon} \operatorname{KL} (f_S  (\mathbf{x}+\boldsymbol{\delta'}), f_T  (\mathbf{x})), \\
    \text{AdaAD attack: } & \boldsymbol{\delta} = \argmax_{ \| \boldsymbol{\delta'} \|_p \leq \epsilon} \operatorname{KL} (f_S  (\mathbf{x}+\boldsymbol{\delta'}), f_T  (\mathbf{x} +\boldsymbol{\delta'})).
\end{align}
Here, $\text{CE}$ represents cross-entropy loss, $\text{KL}$ represents KL-divergence loss.

\subsection{Hyperparameter}
The parameters of AT, AD, and the AD component in IGDM-incorporated methods were strictly set to the value suggested in the original paper.
On the other hand, a parameter of IGDM varied depending on the original AT or AD method, dataset, teacher model, and student model.
We experimentally employed an increasing hyperparameter function, defined as $T(\alpha) = \frac{\text{Current Epoch}}{\text{Total Epochs}}\cdot \alpha$, and the $\alpha$ value varied in each training scenario.
The following paragraphs provide detailed information on the $\alpha$ in each experimental setting, determined through grid search results.
IGDM$_{\text{AdaAD}}$ and IGDM$_{\text{AdaIAD}}$ have same value of hyperparameter in all cases.

\subsubsection{ResNet-18 model trained on CIFAR-100 dataset}
For IGDM$_{\text{PGD-AT}}$, to 100 for LTD and IKL-AT, 220 for BDM-AT.
For IGDM$_{\text{TRADES}}$, 5 for all teachers.
For IGDM$_{\text{ARD}}$, 50 for LTD and 100 for BDM-AT and IKL-AT.
For IGDM$_{\text{IAD}}$, 20 for LTD and 50 for BDM-AT and IKL-AT.
For IGDM$_{\text{AKD}}$, 50 for LTD, 25 for BDM-AT, and 40 for IKL-AT.
For IGDM$_{\text{RSLAD}}$, 3 for LTD and 10 for BDM-AT and IKL-AT.
For IGDM$_{\text{AdaAD}}$, 15 for LTD, 70 for BDM-AT, and 50 for IKL-AT.

\subsubsection{ResNet-18 model trained on the CIFAR-10 dataset}
For IGDM$_{\text{PGD-AT}}$, 60 for LTD and 70 for BDM-AT and IKL-AT.
For IGDM$_{\text{TRADES}}$, 20 for all teachers.
For IGDM$_{\text{ARD}}$, 5 for LTD and 10 for BDM-AT and IKL-AT.
For IGDM$_{\text{IAD}}$, 50 for LTD and 30 for BDM-AT and IKL-AT.
For IGDM$_{\text{AKD}}$, 7 for LTD, 10 for BDM-AT, and 15 for IKL-AT.
For IGDM$_{\text{RSLAD}}$, 0.9 for LTD and BDM-AT, and 1 for IKL-AT.
For IGDM$_{\text{AdaAD}}$, 5 for LTD and IKL-AT, and 10 for BDM-AT.

\subsubsection{MobileNetV2 model trained on CIFAR-100 dataset}
For IGDM$_{\text{PGD-AT}}$, 150 for LTD, 160 for BDM-AT, and IKL-AT.
For IGDM$_{\text{TRADES}}$, 10 for LTD and BDM-AT, and 3 for IKL-AT.
For IGDM$_{\text{ARD}}$, 70 for all teachers.
For IGDM$_{\text{IAD}}$, 20 for LTD and 50 for BDM-AT and IKL-AT.
For IGDM$_{\text{AKD}}$, 30 for all teachers.
For IGDM$_{\text{RSLAD}}$, 3 for LTD, 4 for BDM-AT, and 10 for IKL-AT.
For IGDM$_{\text{AdaAD}}$, 20 for LTD, 50 for BDM-AT, and 40 for IKL-AT.

\subsubsection{MobileNetV2 model trained on CIFAR-10 dataset}
For IGDM$_{\text{PGD-AT}}$, 70 for LTD, 40 for BDM-AT, and 50 for IKL-AT.
For IGDM$_{\text{TRADES}}$, 1 for LTD, 0.5 for BDM-AT, and 0.2 for IKL-AT.
For IGDM$_{\text{ARD}}$, 10 for LTD, 8 for BDM-AT, and 9 for IKL-AT.
For IGDM$_{\text{IAD}}$, 50 for all teachers.
For IGDM$_{\text{AKD}}$, 15 for all teachers.
For IGDM$_{\text{RSLAD}}$, 0.5 for LTD and 0.7 for BDM-AT and IKL-AT.
For IGDM$_{\text{AdaAD}}$, 10 for all teachers.

\subsubsection{ResNet-18 model trained on the SVHN dataset}
We set 60 for IGDM$_{\text{PGD-AT}}$, 30 for IGDM$_{\text{TRADES}}$, 50 for IGDM$_{\text{ARD}}$, 40 for IGDM$_{\text{IAD}}$, 10 for IGDM$_{\text{AKD}}$, 3 for IGDM$_{\text{RSLAD}}$, and 9 for IGDM$_{\text{AdaAD}}$.

\subsubsection{PreActResNet-18 model trained on Tiny-ImageNet dataset}
We set 30 for IGDM$_{\text{PGD-AT}}$, 20 for IGDM$_{\text{TRADES}}$, 10 for IGDM$_{\text{ARD}}$, 20 for IGDM$_{\text{IAD}}$, 10 for IGDM$_{\text{AKD}}$, 1 for IGDM$_{\text{RSLAD}}$, and 10 for IGDM$_{\text{AdaAD}}$.

%\subsection{Logit Adjustment}

\section{Main Algorithm}

\begin{algorithm}[h]
\caption{Main Algorithm}
\label{alg:algorithm}
\hspace*{0.1in}\textbf{Input:} Robust teacher model $f_T$, inner loss ($L_{max}$) and outer loss ($L_{AD}$) of base AT 
or AD \hspace*{0.2in}method, and batched training dataset $\{(\mathbf{x}, y)\}$ with $n$ batch size.\\
\hspace*{0.1in}\textbf{Output:} Robust stduent model $f_S$ \vspace{0.1in}\\
\hspace*{0.1in}Randomly initialize $\theta$, the weights of $f_{S}$ \\
\hspace*{0.1in}\textbf{repeat} 
\\
\hspace*{0.2in}$\boldsymbol{\delta} \ = \ \underset{\| \boldsymbol{\delta'} \|_\infty \leq \epsilon}{\argmax}\left   (L_{max}(\mathbf{x}+\boldsymbol{\delta'}, y ) \right )$ 
\vspace{0.05in}\\
\hspace*{0.2in}$L_{IGDM}(\mathbf{x}, \boldsymbol{\delta}) = T(\alpha) \cdot D(f_S(\textbf{x} +  \boldsymbol{\delta} ) - f_S(\textbf{x} -  \boldsymbol{\delta} ), \ f_T(\textbf{x} +  \boldsymbol{\delta}) - f_T(\textbf{x} -  \boldsymbol{\delta} ))$
\vspace{0.05in}\\
\hspace*{0.2in}$L_{min}(\mathbf{x}, y) = L_{AD}(\mathbf{x}, y) + L_{IGDM}(\mathbf{x}, \boldsymbol{\delta})$
\vspace{0.05in}\\
\hspace*{0.2in}$\theta \gets -\eta \frac{1}{n} \sum_{i=1}^n \nabla_{\theta} L_{min}(\mathbf{x}_i, y_i)$
\vspace{0.05in}\\
\hspace*{0.1in}\textbf{until} training converged

\end{algorithm}

\end{document}